\pdfoutput=1
\documentclass[preprint,12pt]{elsarticle}

\usepackage{amssymb}
\usepackage{amsmath}
\usepackage{algorithm}
\usepackage{algorithmic}
\usepackage{graphicx}

\usepackage{booktabs}
\usepackage{multirow}

\usepackage{color}
\newcommand{\red}[1]{\textcolor{red}{#1}}

\journal{Pattern Recognition}

\begin{document}

\begin{frontmatter}



\title{Class-agnostic 3D Segmentation by Granularity-Consistent Automatic 2D Mask Tracking} 


\author[label1]{Juan~Wang}
\author[label2]{Yasutomo~Kawanishi}
\author[label1]{Tomo~Miyazaki}
\author[label3]{Zhijie~Wang}
\author[label1]{Shinichiro~Omachi}

\affiliation[label1]{organization={Graduate School of Engineering},
            addressline={Tohoku University}, 
            city={Sendai},
            country={Japan}}\

\affiliation[label2]{organization={Multimodal Data Recognition Research Team},
            addressline={RIKEN GRP}, 
            city=Kyoto,
            country={Japan}}\
            
\affiliation[label3]{organization={Multimodal Visual Intelligence Team},
            addressline={RIKEN AIP}, 
            city=Sendai,
            country={Japan}}




\begin{abstract}

3D instance segmentation is an important task for real-world applications. To avoid costly manual annotations, existing methods have explored generating pseudo labels by transferring 2D masks from foundation models to 3D. However, this approach is often suboptimal since the video frames are processed independently. This causes inconsistent segmentation granularity and conflicting 3D pseudo labels, which degrades the accuracy of final segmentation. To address this, we introduce a Granularity-Consistent automatic 2D Mask Tracking approach that maintains temporal correspondences across frames, eliminating conflicting pseudo labels. Combined with a three-stage curriculum learning framework, our approach progressively trains from fragmented single-view data to unified multi-view annotations, ultimately globally coherent full-scene supervision. This structured learning pipeline enables the model to progressively expose to pseudo-labels of increasing consistency. Thus, we can robustly distill a consistent 3D representation from initially fragmented and contradictory 2D priors. Experimental results demonstrated that our method effectively generated consistent and accurate 3D segmentations. Furthermore, the proposed method achieved state-of-the-art results on standard benchmarks and open-vocabulary ability.
\end{abstract}

\let\newpage\relax
\let\clearpage\relax
\let\cleardoublepage\relax


\end{frontmatter}



\section{Introduction}
\label{sec1}

3D instance segmentation is a fundamental task in computer vision and robotics, which aims at partitioning 3D scenes into semantically meaningful segments at the instance level. Current fully supervised instance segmentation methods have achieved significant progress and can generate high-quality 3D proposals, such as Mask3D~\citep{schult2023mask3d} and SoftGroup~\citep{vu2022softgroup}. However, these approaches require annotated datasets for training, presenting two major limitations. Firstly, manual 3D annotation is expensive and time-consuming. Secondly, these methods are confined to a narrow range of object categories within specific closed-set 3D datasets, such as ScanNet~\citep{dai2017scannet}, ScanNet200~\citep{rozenberszki2022language-scannet200}, and ScanNet++~\citep{yeshwanth2023scannet++}. The limitations greatly restricts their real-world applications in domains like embodied agents and autonomous driving.

\begin{figure*}[!t]
\centering
\includegraphics[width=\textwidth]{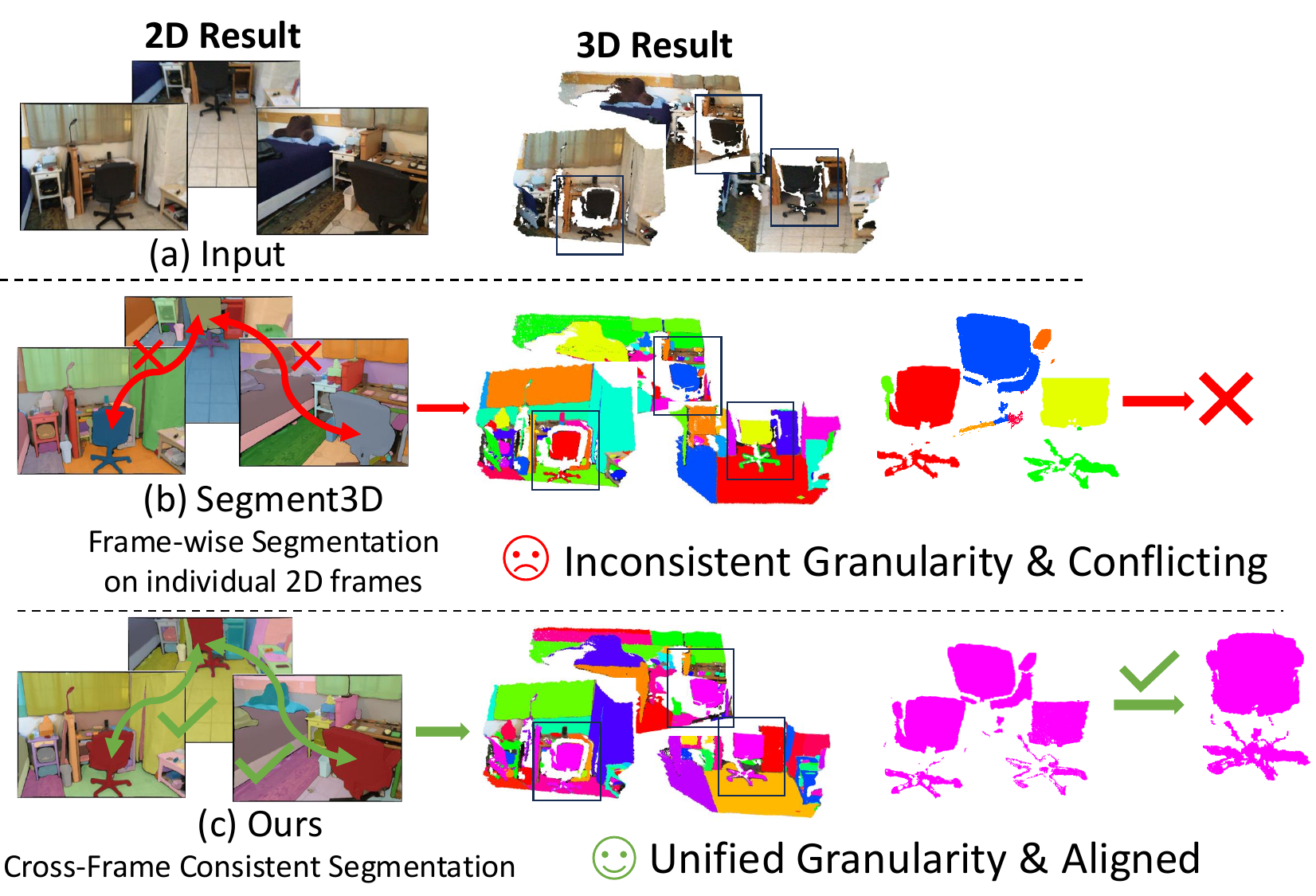}
\caption{Comparison of pseudo label generation between our method and existing class-agnostic 3D instance segmentation approach Segment3D~\cite{huang2024segment3d}. (a) Input RGB-D video frames from an indoor scene in ScanNet~\citep{dai2017scannet}, showing the same chair object from different viewpoints.
(b) Segment3D~\cite{huang2024segment3d} employs Automatic-SAM~\cite{kirillov2023segment} on individual 2D frames to generate frame-specific masks, resulting in inconsistent segmentation granularity. For example, the chair object is segmented with different levels of detail across frames, producing conflicting pseudo-labels in 3D space. 
(c) Our method incorporates a cross-frame consistent segmentation module that maintains object tracking across 2D frames, ensuring consistent segmentation granularity throughout the video sequence. This produces masks with unified segmentation boundaries across frames, leading to coherent 3D pseudo-labels. For example, in the 3D results shown for the chair, our method generates unified results.}\label{fig:comparison}
\end{figure*}

Recent open-vocabulary methods have explored class-agnostic instance segmentation to overcome these limitations. Training-free methods~\citep{lu2023ovir,yang2023sam3d,yin2024sai3d} lift 2D masks to 3D through multistage pipelines but suffer from manual parameter tuning, error accumulation, and slow inference times of several minutes per scene. In contrast, training-based methods~\citep{qi2022high-cropformer,guo2024sam-graph} use pseudo-labels from 2D foundation models such as SAM~\citep{kirillov2023segment} to train 3D segmenters. This approach enables real-time inference while avoiding error accumulation and reducing manual parameter tuning. However, the existing methods such as Segment3D~\citep{huang2024segment3d} perform frame-individual segmentation without considering inter-frame associations, leading to inconsistent 2D masks and conflicting 3D pseudo labels.

To address this problem, we propose a Granularity-Consistent Segmentation Policy combined with a three-stage curriculum learning framework. First, as shown in Fig.~\ref{fig:comparison}(c), our method establishes temporal correspondences across video frames by automatically tracking objects, generating 2D masks with consistent granularity across frames. This tracking mechanism resolves the fragmentation inconsistencies illustrated in Fig.~\ref{fig:comparison}(b), where the same object is fragmented differently between frames. This produces unified pseudo-labels with consistent granularity for 3D Projection. Second, we propose a three-stage curriculum learning pipeline, as illustrated in Fig.~\ref{fig:overview}, based on these consistent annotations. This pipeline progressively exposes the model to pseudo-labels of increasing consistency and completeness. In Stage 1, the model learns from fragmented single-view information extracted from key frames. In Stage 2, the model is trained using temporally consistent annotations with uniform segmentation granularity generated by our tracking policy, enabling it to learn robust cross-view correspondences and temporal relationships. In Stage 3, we fine-tune the model on complete scene point clouds to enforce global geometric coherence across the entire scene. We evaluated our model on ScanNet200~\citep{rozenberszki2022language} and ScanNet++~\citep{yeshwanth2023scannet++}, achieving state-of-the-art results. Our contributions are summarized as follows:\par
\begin{itemize}
    \item We design a Granularity-Consistent Segmentation Policy to establish temporal correspondences across frames via automatically 2D Mask Tracking, generating coherent 3D pseudo-labels across multi views.
    \item We introduce a three-stage curriculum learning framework that progressively trains the model from fragmented single-view data, through consistent multi-view annotations, to full-scene supervision enabling robust 3D segmentation learning from initially conflicting 2D priors.
    \item We demonstrated that the generated 3D pseudo-labels were more accurate than the existing methods with extensive experiments. Also, the results verified the importance of consistency and the generalization ability of our method.
    \item We validated our method's open-vocabulary capabilities through text-based retrieval, demonstrating superior performance in fine-grained object recognition and long-tail categories-rare objects with limited training samples, especially for out-of-vocabulary queries.
\end{itemize}

\section{Related works}
\label{sec2}
We review existing works that are relevant to ours, including fully-supervised 3D instance segmentation (Sec.~\ref{subsec2-1}), class-agnostic 3D instance segmentation (Sec.~\ref{subsec2-2}), the Segment Anything Model (Sec.~\ref{subsec2-3}), and Open-Vocabulary 3D Scene Understanding (Sec.~\ref{subsec2-4}).

\subsection{Fully-Supervised 3D Instance Segmentation}
\label{subsec2-1}
Fully-supervised 3D instance segmentation~\citep{wang2023openinst, qi2022open, shen2021high, qi2022cassl, qi2022high-cropformer} aims to identify and segment individual object instances in 3D scenes using models trained on datasets with complete class labels and instance annotations. The field has evolved from proposal-based~\citep{yang2019learning, yi2019gspn} and grouping-based~\citep{chen2021hierarchical, jiang2020pointgroup, liang2021instance, vu2022softgroup} approaches to recent transformer-based~\citep{schult2023mask3d, lu2023query, wu2024ppt, wu2022ptv2, wu2024ptv3, wu2024ppt} architectures. Among these methods, Mask3D~\citep{schult2023mask3d} has emerged as a representative approach, employing a transformer-based architecture with learned queries to predict instance masks. These methods achieve high-quality segmentation on benchmark datasets through supervised learning on closed-set annotations.

The fully-supervised methods face significant limitations: they require extensive manually labeled training data, which is expensive and time-consuming to obtain, and they can only recognize predefined categories within specific datasets such as ScanNet~\citep{dai2017scannet} and ScanNet200~\citep{rozenberszki2022language-scannet200}, severely restricting their applications in open-world scenarios. 

We focus on the architectural designs of fully-supervised methods because of their geometric reasoning capabilities. Therefore, we adopt Mask3D's transformer-based architecture as our backbone network. Then we fundamentally modify its training paradigm. Specifically, we train it using automatically generated pseudo-labels from 2D foundation models instead of relying on manual closed-set annotations. Consequently, the proposed training paradigm enables class-agnostic segmentation that generalizes beyond predefined vocabularies.


\subsection{Class-Agnostic 3D Instance Segmentation}
\label{subsec2-2}

Class-agnostic 3D instance segmentation aims to detect and segment object instances without predefined class labels. This approach decouples geometric segmentation from semantic classification~\citep{wu2025class} to address the scalability limitations of fully-supervised methods. Existing approaches can be categorized into two strategies: training-free and training-based. 

Training-free methods~\citep{lu2023ovir,yang2023sam3d,yin2024sai3d} typically transform point clouds into superpoints using hand-crafted algorithms. Then, 2D masks from foundation models are projected into 3D space and fused in a bottom-up manner to generate superpoints. While avoiding the need for training, these methods have several limitations. They require manually designed fusion strategies and lack adaptive learning capability. Additionally, error accumulation across multiple processing stages is unavoidable. Moreover, their inference time of several minutes per scene makes them unsuitable for real-time applications.

Training-based methods~\citep{qi2022high-cropformer,guo2024sam-graph} leverage pseudo-labels obtained by 2D foundation models, such as SAM~\citep{kirillov2023segment}, to train 3D segmenters. The approach enables real-time inference and end-to-end learning, effectively mitigating error accumulation and reducing reliance on manual parameter tuning. However, even a state-of-the-art method~\citep{huang2024segment3d} needs to process each frame independently, resulting in conflicting 2D masks and temporally inconsistent 3D pseudo-labels. As illustrated in Fig.~\ref{fig:comparison}(b), independent frame processing causes the same object to be segmented with varying granularity across frames, creating contradictory pseudo-labels when projected to 3D space.

To address the problem of inconsistency of pseudo-labels, we propose a cross-frame consistent segmentation approach that establishes temporal correspondences across video frames through automatic 2D mask tracking. The proposed method ensures consistent segmentation granularity throughout sequences. As shown in Fig.~\ref{fig:comparison}(c), this produces unified pseudo-labels across multiple views, providing higher-quality supervision for training the 3D segmentation network.

\subsection{The Segment Anything Model (SAM)}
\label{subsec2-3}
The Segment Anything Model (SAM)~\cite{kirillov2023segment} has revolutionized 2D segmentation by enabling zero-shot segmentation of arbitrary objects. The existing methods, Segment3D~\cite{huang2024segment3d}, SAI3D~\cite{yin2024sai3d}, and SAM3D~\cite{yang2023sam3d}, apply SAM's automatic mask generation to individual frames for 3D tasks. Unfortunately, by treating frames independently, they suffer from the same temporal inconsistency issues, yielding conflicting segmentations and suboptimal 3D results. The recent introduction of SAM2~\cite{ravi2024sam2} incorporates temporal tracking for video. Motivated by this, we propose a granularity-consistent segmentation policy that leverages SAM's precise single-frame automatic segmentation capability for keyframe detection, while utilizing SAM2's temporal propagation mechanism to maintain cross-frame correspondences. By integrating object state management to coordinate between keyframe detection and temporal tracking, our approach generates temporally consistent 2D masks that resolve inter-frame conflicts and enable coherent 3D scene understanding.


\subsection{Open-Vocabulary 3D Scene Understanding}
\label{subsec2-4}
Open-vocabulary 3D scene understanding~\citep{zhang2023clipfo3d, ha2022semabs, sun2025ov, zhang2025clip-pr} aims to recognize and segment objects using arbitrary text descriptions. Vision-language models, such as CLIP~\cite{radford2021learning-clip}, are used to enable open-world perception beyond closed-set categories. Broadly, there are semantic- and instance-level approaches.

The semantic-level methods focus on point-level open-vocabulary recognition. Related works such as PLA~\citep{ding2022language-pla} and RegionPLC~\citep{yang2024regionplc} aligns point cloud features with captions extracted from multi-view images to enable open-vocabulary understanding. OpenScene~\citep{Peng2023OpenScene} distills per-pixel CLIP features from 2D images to 3D point clouds, creating point-wise representations co-embedded with text in CLIP feature space. While these methods achieve open-vocabulary semantic segmentation, they primarily operate at the point or region level and exhibit limited capability in distinguishing individual object instances, which is essential for tasks requiring precise object-level understanding and manipulation.
 
The instance-level approaches address open-vocabulary understanding by combining geometric instance segmentation with vision-language features. One approach employs 2D segmentation models to generate view-specific masks that are lifted into 3D space, exemplified by SAI3D~\citep{yin2024sai3d} and OVIR-3D~\citep{lu2023ovir}. While benefiting from rich semantic information for detecting small objects, these methods often struggle with temporal and geometric consistency when aggregating masks across views. Alternatively, OpenMask3D~\citep{takmaz2023openmask3d} uses fully-supervised 3D segmenter to predict class-agnostic masks. Then, the 3D masks are associated to CLIP features through multi-view aggregation for open-vocabulary retrieval. 

However, reliance on closed-set training annotations potentially limits generalization to novel categories. Thus, our method bridges these approaches by generating class-agnostic proposals through temporally consistent 2D mask tracking, eliminating the need for manual annotations. This design address both the temporal inconsistency of frame-independent methods and the closed-set limitations of supervised approaches. To validate open-vocabulary capabilities, we adopt OpenMask3D's~\citep{takmaz2023openmask3d} multi-view feature aggregation protocol (Sec.~\ref{sec:Open_voca}) and demonstrate effectiveness on both standard benchmark categories and out-of-vocabulary queries.

\section{Method}
\label{method}

\begin{figure*}[!t]
    \centering
    \includegraphics[width=\textwidth]{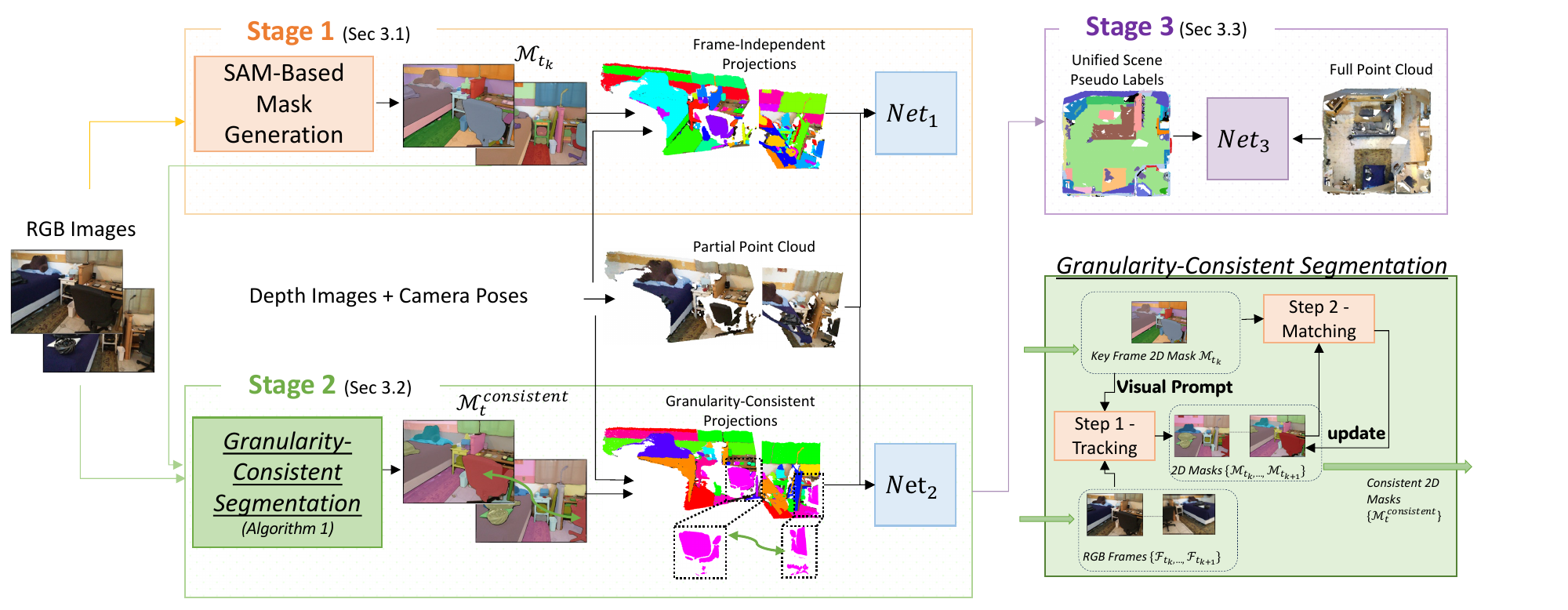}
    \caption{An overview of the proposed method. We propose a Granularity-Consistent Segmentation Policy with three-stage curriculum learning pipeline for class-agnostic 3D instance segmentation. Stage 1: From input RGB-D video sequences, we apply SAM-Based Mask Generation to extract initial 2D masks $\mathcal{M}_{t_k}$ on keyframes ${t_k}$, which are then projected to 3D space as frame-independent pseudo labels for fragmented warm-up training of model $\text{Net}_1$. Stage 2: Our Granularity-consistent Segmentation Policy generates 2D Mask $\mathcal{M}_{t}^{consistent}$ and projected as Granularity-Consistent pseudo labels across all frames to finetune and obtain model $\text{Net}_2$. Stage 3: We fine-tune the model on full point clouds with confidence-based filtering to achieve globally coherent class-agnostic 3D instance segmentation, yielding the final model $\text{Net}_3$.}
    
    \label{fig:overview}
\end{figure*}

As illustrated in Fig.~\ref{fig:comparison}(b), the existing method~\citep{huang2024segment3d} produced masks with inconsistent granularity, resulting in conflicting 3D pseudo-labels. To address this, we propose a Granularity-Consistent Segmentation Policy (Sec.~\ref{sec:stage2}) that maintains temporal correspondences through automatic 2D mask tracking, ensuring unified segmentation boundaries across frames. Building upon this policy, we design a three-stage curriculum learning pipeline shown in Fig.~\ref{fig:overview}. We progressively expose the model to pseudo-labels of increasing consistency and completeness. This structured progression is essential because a direct training on high-quality consistent annotations is difficult without proper initialization. Therefore, we train the model by gradual exposure, which is started from fragmented single-view data to consistent multi-view annotations, and finally to full-scene supervision.

The Stage 1 is fragmented warm-up training. We obtain 2D segmentation results on individual key frames independently. Then, we project the masks into 3D space to serve as pseudo labels for initial model training. Despite producing frame-wise inconsistent segmentation, the Stage 1 establishes basic object-level feature representations necessary for subsequent refinement. The Stage 2 is granularity-consistent segmentation learning. We propose a granularity-consistent segmentation policy to generate temporally consistent 2D masks across all frames through tracking and matching. Then, we transform the 2D masks into unified 3D pseudo-labels for fine-tuning the model. Thus, we can alleviate the cross-frame granularity inconsistencies from Stage 1. Hence, we enable learning of robust cross-view correspondences. The Stage 3 is full-scene fine-tuning. We further fine-tune on complete 3D point clouds to enforce global geometric coherence, transitioning from partial-view understanding to holistic scene reasoning.

For each scene, we have video sequence $\mathcal{V}=\{F_t\}_{t=1}^{T}$ including $T$ frames, with corresponding depth images $\{D_{t}\}_{t=1}^{T}$, where $D_{t} \in \mathbb{R}^{H \times W}$, $H$ and $W$ are height and width, respectively. $A\in \mathbb{R}^{3 \times 3}$ denotes the camera intrinsic matrix, and $\{E_{t}\}_{t=1}^{T}$ denotes a set of camera extrinsic matrices. In addition, complete 3D point cloud $\mathcal{P}_\mathrm{full} = \{p_i\}_{i=1}^{N}$, where each point $p_i \in \mathbb{R}^3$ represents a 3D coordinate in the scene, is also provided by the dataset.

\subsection{Stage 1: Fragmented Warm-up Training}
\label{sec:stage1}
At this stage, the model undergoes warm-up training on fragmented pseudo labels. We generate 2D masks on keyframes and project them into 3D space to obtain pseudo labels for training the 3D segmentation model.

\subsubsection{SAM-Based 2D Mask Generation}
Processing every frame of the video sequence $\mathcal{V}$ is computationally expensive and introduces significant redundancy since adjacent frames are often highly correlated. Extracting some frames to represent the entire sequence can achieve a good balance between computation and performance. Therefore, we sampled a set of keyframes $\mathcal{K} = \{F_{t_k}\}_{k=1}^{K}$ from $\mathcal{V}$ with stride $s$, along with corresponding depth images, camera intrinsic matrix, and extrinsic matrix. 

We apply SAM's automatic mask generation~\citep{kirillov2023segment} to each keyframe $F_{t_k}$. SAM employs a multi-scale strategy that processes both the full image and multiple cropped regions at different resolutions, producing binary 2D masks at various granularities, where pixel values of 0 and 1 indicate background and object instance, respectively. However, this multi-scale processing introduces redundant detections where the same object is segmented differently across scales. For instance, consider a chair object: SAM may generate a complete mask covering the entire chair at the full-image scale, while simultaneously producing separate masks for the seat, backrest, and legs at finer crop scales. These fine-grained component masks are substantially contained within the coarse-grained complete mask, creating redundancy that needs be eliminated. To identify such redundant masks, denoted as $\mathcal{M}_{\text{redundant}}$, we employ a containment-based filtering strategy following Segment3D~\citep{huang2024segment3d}:
\begin{equation}
\mathcal{M}_{\text{redundant}} = \left\{M \in \mathcal{M}_{\text{fine}} \mid 
\exists M' \in \mathcal{M}_{\text{coarse}}, 
\frac{|M \cap M'|}{|M|} > \tau_{\text{contain}}\right\}
\label{eq:contained_masks}
\end{equation}
where $\mathcal{M}_{\text{fine}}$ represents fine-grained masks from full-resolution processing, and $\mathcal{M}_{\text{coarse}}$ contains coarse-grained masks from crop-based processing at larger scales. The ratio of $\frac{|M \cap M'|}{|M|}$ measures the \textit{containment rate} $\tau_{contain}$, the proportion of mask $M$ that is covered by the larger mask $M'$. We use containment rate rather than standard Intersection over Union (IoU) because standard IoU (intersection/union) is influenced by the area of the larger mask, whereas containment rate (intersection/area of small mask) directly quantifies the degree to which the small mask is covered. We set the containment threshold $\tau_{\text{contain}} = 0.8$, such that masks with over 80\% overlap are considered redundant. 

Having identified $\mathcal{M}_{\text{redundant}}$, we remove these redundant masks from SAM's output to eliminate conflicting segmentations. The final mask set for each keyframe is obtained as:
\begin{equation}
 \mathcal{M}_{t_k}^{Stage1} = \{M_{t_k}^{(i)}\}_{i=1}^{N_{t_k}} =  \text{SAM}(F_{t_k}) \setminus \mathcal{M}_{redundant}
\label{eq:stage1_2d_mask}
\end{equation}
where $M_{t_k}^{(i)} \in \{0,1\}^{H \times W}$ represents the binary mask for the $i$-th object, and $N_{t_k}$ is the total number of non-redundant objects detected in keyframe $F_{t_k}$. The set difference operator $\setminus$ denotes removal of redundant masks. This post-processing yields high-quality, non-redundant binary 2D masks per keyframe, effectively reducing the number of instances and mitigating conflicts for subsequent 3D projection.


\subsubsection{3D Mask Preparation}
For each 2D mask $M_{t_k}^{(i)}$, we project it to 3D space using the corresponding depth image and camera parameters. We obtain the point cloud set of $i$-th object through a two-step transformation. Firstly, we transform pixels to camera coordinates by Eq.~\eqref{eq:pixel_to_camera}. $A = [ f_x, 0, c_x ; 0, f_y, c_y ; 0, 0, 1 ]$ is the camera intrinsic matrix with focal lengths $f_x, f_y$ and principal point $(c_x, c_y)$, $D_{t_k}(u,v)$ is the depth value at pixel $(u,v)$. Secondly, we transform camera to world coordinates by Eq.~\eqref{eq:camera_to_world}. $E_{t_k}$ is the camera pose matrix of the $t_k$-th frame. 
Therefore, a point cloud set $\mathcal{P}_{t_k}$ for the $i$-th object is defined as Eq.~\eqref{eq:stage1_point_set}, where $(u,v)$ are pixel coordinates in the 2D mask, and $(X_w, Y_w, Z_w)$ represents the corresponding 3D world coordinates.
\begin{align}
    [X, Y, Z]^T &= D_{t_k}(u,v) \cdot A^{-1} [ u, v, 1 ]^T, \label{eq:pixel_to_camera} \\
    [X_w, Y_w, Z_w, 1]^T &= E_{t_k} \cdot [X, Y, Z, 1]^T, \label{eq:camera_to_world} \\
    \mathcal{P}_{t_k}^{(i)} &= \{(X_w, Y_w, Z_w) | (u,v) \in M_{t_k}^{(i)}, D_{t_k}(u,v) > 0\}.
\label{eq:stage1_point_set}
\end{align}
where $P_{t_k}^{(i)}$ denotes the 3D point set corresponding to the $i$-th 2D mask $M_{t_k}^{(i)}$ from Eq. (1). We assign instance label $i$ to all points in $P_{t_k}^{(i)}$, thereby establishing the correspondence between 2D masks $\mathcal{M}^{\text{Stage1}}$ and 3D pseudo-labels $\tilde{Y}^{(1)}$.



We discard point sets if they are fewer than 100 points. The remaining 3D pseudo-labels are aggregated across all keyframes for warm-up training.
\begin{equation}
    \tilde{Y}^{(1)} = \bigcup_{k=1}^{K} \bigcup_{i=1}^{N_{t_k}} \mathcal{P}_{t_k}^{(i)}.
\end{equation}
where $\bigcup$ denotes set union operation, the superscript $(1)$ indicates Stage 1 and $\tilde{Y}^{(1)}$ are the 3D pseudo-labels of Stage 1. By this way, we create keyframes's partial 3D point clouds and their corresponding 3D pseudo labels as the training dataset and annotations.

\subsubsection{Model Training} 

The architecture of the proposed segmentation model is based on a query-based segmentation framework inspired by Mask3D~\cite{schult2023mask3d}, consisting of a MinkowskiUNet~\cite{choy20194d} backbone and a transformer decoder for sparse point cloud extraction and instance prediction, respectively. The decoder initializes a set of learnable queries using furthest point sampling and Fourier encodings. Through cross-attention mechanisms, these queries iteratively interact with multi-scale point features to generate both mask embeddings and objectness scores. Final instance masks are produced by computing similarity between query embeddings and point features, followed by thresholding to obtain binary segmentation results.

We marked the segmentation model at Stage 1 as $\text{Net}_1$ and train it using the generated 3D pseudo-labels with the following objective function: 
\begin{equation}
\begin{aligned}
\mathcal{L}_\mathrm{stage1} ={} & \sum_{p \in \tilde{Y}^{(1)}} \left[ \lambda_\mathrm{dice} \mathcal{L}_\mathrm{dice}(\text{Net}_1(p), \hat{y}_p) + \lambda_\mathrm{ce} \mathcal{L}_\mathrm{ce}(\text{Net}_1(p), \hat{y}_p) \right] \\
                       & + \lambda_\mathrm{obj} \mathcal{L}_\mathrm{obj}.
\label{eq:stage1_loff_function}
\end{aligned}
\end{equation}
where $\hat{y}_p$ is the pseudo-label for point $p$ and $\text{Net}_1(p)$ is corresponding prediction result. The training objective combines three components: $\mathcal{L}_\mathrm{dice}$ is the Dice loss that measures overlap between predicted and ground truth masks, providing robust optimization for segmentation boundaries; $\mathcal{L}_\mathrm{ce}$ is the cross-entropy loss that enforces pixel-wise classification accuracy; and $\mathcal{L}_\mathrm{obj}$ is the objectness loss that predicts the confidence score for object existence. The hyperparameters $\lambda_\mathrm{dice}$, $\lambda_\mathrm{ce}$, and $\lambda_\mathrm{obj}$ balance the contributions of each loss component during training.

\subsection{Stage 2: Granularity-Consistent Segmentation Learning}
\label{sec:stage2}
In Stage 2, we leverage SAM2~\citep{ravi2024sam2} mask propagation capability to not only transmit mask results from previous key frames to subsequent frames but also capture different views of the same object, such as the chair shown at various views in Fig.~\ref{fig:comparison}. Additionally, we propose an object robust status management mechanism that applies for objects potentially disappearing in intermediate frames and reappearing in later frames. Since SAM2 only tracks within a limited temporal window, we assign a status \textit{dormant} to temporarily disappeared objects. When such objects reappear in later frames, we compare their IoU with dormant objects, enabling the system to maintain the ability to recognize and track temporarily occluded objects that later become visible again. 
\begin{algorithm}[!ht]
\caption{Cross-Frame Consistent Segmentation}
\label{alg:consistent_seg_alg}
\textbf{Input:} Video $\mathcal{V} = \{F_t\}_{t=1}^T$, stride $s$, threshold $\tau_{IoU}$  \\
\textbf{Output:} Consistent 2D Masks $\mathcal{M}_t^{consistent}$
\begin{algorithmic}[1]
\STATE \textbf{Initialize:} $\mathcal{T}$, $\mathcal{P}$ \COMMENT{\textbf{T}racker, \textbf{P}rompt loader}
\FOR{$t_k \in \mathcal{K} = \{s, 2s, ...\}$}
    \STATE $\mathcal{M}_{t_k}^{Stage1}, \mathcal{M}_{t_k}^{Stage2} \leftarrow$ Load masks at keyframe $t_k$ in Eq.~\eqref{eq:stage1_2d_mask}
    \STATE $\mathcal{S} \leftarrow \text{OptimalMatch}(\mathcal{M}_{t_k}^{Stage1}, \mathcal{M}_{t_k}^{Stage2}, \tau_{IoU})$ in Eq.~\eqref{eq:stage2_iou_matching}
    \STATE $\mathcal{T} \leftarrow \text{UpdateStates}(\mathcal{S})$ \COMMENT{Active/Dormant/Terminated} in Fig.~\ref{fig:status_transitions}
    \STATE $\mathcal{P} \leftarrow \text{AddPrompts}(\mathcal{T}.\text{active\_objects}, t_k)$
    \STATE $\mathcal{M}_{[t_k, t_{k+s}]}^{consistent} \leftarrow \text{SAM2\_Propagate}(\mathcal{P})$
\ENDFOR
\RETURN $\bigcup_{t_k} \mathcal{M}_{[t_k, t_{k+1}]}^{consistent}$
\end{algorithmic}
\end{algorithm}

\begin{figure*}[!t]
    \centering
    \includegraphics[width=.6\textwidth]{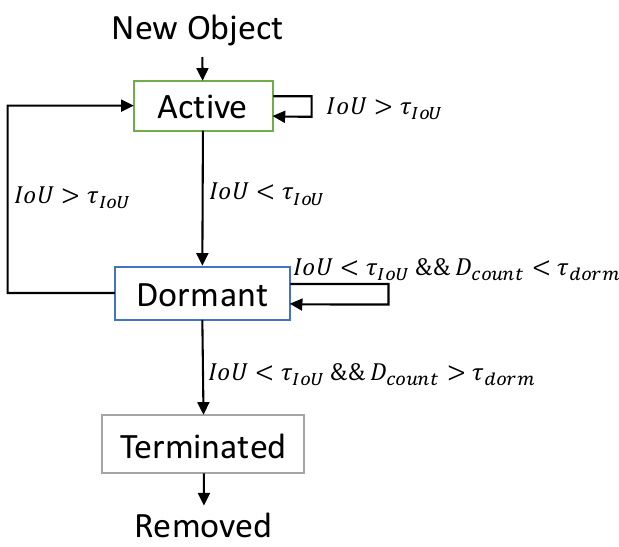}
    \caption{Object Status Transitions. Our state management system handles three object states: Active (currently tracked), Dormant (temporarily lost), and Terminated (permanently removed). Transitions are governed by IoU matching thresholds $\tau_{IoU}$, dormancy counters $D_{count}$, and dormancy threshold $\tau_{dorm}$, enabling robust tracking across temporary occlusions and viewpoint changes.}
    \label{fig:status_transitions}
\end{figure*}

\subsubsection{Granularity-Consistent Segmentation via 2D Mask Trcking} 
In this stage, our objective is to establish cross-frame object tracking and ensure temporally granularity consistency of segmentation through inter-frame relationships. To understand the progress, we first provide a quick review of The Segment Anything Model 2 (SAM2)~\cite{ravi2024sam2}. SAM2 is a transformer-based model trained on the large-scale SA-V video dataset~\cite{ravi2024sam2}, giving it a strong zero-shot capability to track and segment novel objects. Unlike its predecessor~\citep{kirillov2023segment}, SAM2 requires a visual prompt (a point, mask, or bounding box) to initiate segmentation on a video frame sequence. The prompt's quality is crucial for the final tracking performance. In our method, we use the mask generated in Stage 1 as the propagation prompt for SAM2.\par
As shown in Algorithm~\ref{alg:consistent_seg_alg}, given a video sequence $\mathcal{V} = \{F_t\}_{t=1}^T$, keyframe stride $s$, and IoU matching threshold $\tau_{IoU}$, we aim to output consistent 2D masks $\mathcal{M}_t^{consistent}$ for all frames. Detailed Algorithm~\ref{alg:consistent_seg_alg} workflow are as follows. \par
\textbf{Step 1.} Initialization and Windowing Strategy. We first initialize the object tracker $\mathcal{T}$ and prompt loader $\mathcal{P}$ (Algorithm~\ref{alg:consistent_seg_alg} Line 1) to manage object states and SAM2 propagation prompts, respectively. The tracker maintains three object collections: active objects $\mathcal{O}_{active}$, dormant objects $\mathcal{O}_{dormant}$, and terminated objects $\mathcal{O}_{terminated}$. Subsequently, we extract keyframe set $\mathcal{K} = \{0, s, 2s, ...\}$ as Stage1, creating overlapping temporal windows. Each window's starting frame simultaneously serves as the ending frame of the previous window, implementing a sliding window processing fashion. This strategy ensures smooth transitions between adjacent temporal segments and serves as the key mechanism for achieving temporal consistency.\par
\textbf{Step 2.} First Window Processing. When $k=0$ in the Algorithm~\ref{alg:consistent_seg_alg}, we execute the following initialization and tracking operations for the first window.
\begin{itemize}
    \item Load Stage 1 Initial Detection Results. First, loading detection masks generated following Eq.~\eqref{eq:stage1_2d_mask} in Stage1, $\mathcal{M}_{t_0}^{Stage1} = \{M_{t_0}^{(i)}\}_{i=1}^{N_{t_0}}$ on the first keyframe, where $N_{t_0}$ represents the number of objects detected in keyframe $t_0$.
    \item State Initialization. All detected objects' states $\mathcal{T}$ are initialized as ``Active'' and added to $\mathcal{P}$ as visual prompts, since all objects in the first frame are considered newly appeared.
    \item First Window Propagation. We use these initial prompts for SAM2 propagation within the first window $[t_0, t_s]$, as shown below:
\end{itemize}

\begin{equation}
\begin{aligned}
    &\mathcal{T}.\text{initialize\_all\_active}(\mathcal{M}_{t_0}^{Stage1}), \\
    &\mathcal{P}.\text{add\_prompts}(\mathcal{T}.\text{active\_objects}, t_0), \\
    &\mathcal{M}_{[t_0, t_s]}^{Stage2} \leftarrow \text{SAM2\_Propagate}(\mathcal{P}, \mathcal{V}_{[t_0, t_s]}).
 \label{eq:stage2_first_window}
\end{aligned}
\end{equation}

\textbf{Step 3.} Subsequent Window Processing. After the tracking process in the first window, we execute the following key tracking and matching steps starting from the second keyframe ($k \geq s$, Algorithm~\ref{alg:consistent_seg_alg} Line 2).\par
\begin{itemize}
\item Dual Mask Acquisition. We load Stage 1 detection results $\mathcal{M}_{t_k}^{Stage1}$ in Eq.~\eqref{eq:stage1_2d_mask} and SAM2 tracking results $\mathcal{M}_{t_k}^{Stage2}$ in Eq.~\eqref{eq:stage2_first_window} for the current keyframe $t_k$ (Algorithm~\ref{alg:consistent_seg_alg} Line 3).\par
\item IoU Optimal Matching. At the current keyframe $t_k$, we match two types of mask results using IoU similarity in Eq.~\eqref{eq:stage2_iou_matching} as follows~(Algorithm~\ref{alg:consistent_seg_alg} Line 4):
\begin{equation}
 \text{Match}(t_{k}) = \{(j, i, \text{IoU}(M_{t_{k}}^{\mathrm{Stage2},(j)}, M_{t_{k}}^{(i)})) | \text{IoU} > \tau_\mathrm{IoU}\}.
 \label{eq:stage2_iou_matching}
\end{equation}
where $k>0$, $(j, i, \text{IoU})$ represents a matching triplet between Stage 2 object $j$ and Stage 1 object $i$ with their IoU score, subject to the constraint that IoU exceeds threshold $\tau_{IoU}$. If $t_k>T$ means the last keyframe, we skip matching process, and adopt SAM2 propagated results directly for final output. This IoU matching strategy significantly enhances segmentation accuracy in long video sequences by correcting Stage 2 tracking drift with Stage 1 high-precision detection results at keyframes. \par

    \item Robust State Management. Each tracked object can exist in one of three states. Active Objects ($\mathcal{O}_{active}$): Objects currently being tracked with consistent IDs across frames. These objects have successful matches between SAM2 tracking results and Stage 1 keyframe detections, maintaining temporal continuity. Dormant Objects ($\mathcal{O}_{dormant}$): Objects that were previously active but temporarily disappeared from SAM2 tracking results. This state accounts for temporary occlusions, camera movement, or brief exits from the field of view. Each dormant object maintains a dormancy counter, which is filtered with hyperparameter $\tau_{dormant}$. Terminated Objects ($\mathcal{O}_{terminated}$): Objects that have been dormant for more than $\tau_{dormant}$ frames, indicating permanent disappearance from the scene. These objects are removed from active tracking but their historical information is preserved. \par

We update all objects' states based on matching results $\mathcal{S} = \text{Match}(t_k)$ in the previous step, which handles object appearance, disappearance, and reappearance scenarios commonly encountered in real-world videos. State transitions follow these rules(Algorithm Line 5, Fig.~\ref{fig:status_transitions}): 
\begin{itemize}
    \item New → Active: When a new object is detected that cannot be associated with any existing dormant object, meaning a newly appeared object begins to be tracked.
    \item Active $\rightarrow$ Active: When IoU matching succeeds at consecutive keyframes ($\text{IoU} < \tau_{IoU}$), meaning the object remains consistently tracked across frames without interruption.
    \item Active $\rightarrow$ Dormant: When IoU matching fails at a keyframe ($\text{IoU} < \tau_{IoU}$), indicating the object temporarily disappears from view due to occlusion or camera movement.
    \item Dormant $\rightarrow$ Dormant: When the object remains unmatched but the dormancy counter has not exceeded the threshold ($\text{IoU} < \tau_{IoU} \&\& D_{count} \leq \tau_{dorm}$), meaning the object stays temporarily invisible with its dormancy counter incrementing.
    \item Dormant $\rightarrow$ Active: When successful re-matching occurs with $\text{IoU} > \tau_{IoU}$, indicating the previously occluded object reappears and tracking resumes.
    \item Dormant $\rightarrow$ Terminated: When dormancy counter exceeds the threshold ($D_{count} > \tau_{dorm}$), meaning the object has been absent for too long and is considered permanently removed from the scene.
    
\end{itemize}


\textbf{Step 4.} Prompt Preparation and Propagation. The masks of the updated active objects are added to the prompt loader $\mathcal{P}$. These prompts are then used by SAM2 to propagate the segmentation through the next temporal window $[t_k, t_{k+s}]$, generating consistent masks for this segment (Algorithm~\ref{alg:consistent_seg_alg}, Lines 6-7):
\begin{equation}
\begin{aligned}
    &\mathcal{P} \leftarrow \text{AddPrompts}(\mathcal{T}.\text{active\_objects}, t_k), \\
    &\mathcal{M}_{[t_k, t_{k+s}]}^\mathrm{consistent} = \text{SAM2\_Propagate}(\mathcal{P}, \mathcal{V}_{[t_k, t_{k+s}]}).
\end{aligned}
\label{eq:stage2_2d_mask_consistent}
\end{equation}

Through our temporal propagation mechanism, the algorithm significantly reduces computational overhead while maintaining tracking quality. This sparse keyframe strategy enables real-time processing capabilities.\par

\textbf{Step 5.} Iterative Processing and Final Output. Repeat the above subsequent window processing until all keyframes are handled. The final output is the union of consistent masks from all windows (Algorithm~\ref{alg:consistent_seg_alg} Line 9): $\bigcup_{t_k} \mathcal{M}_{[t_k, t_{k+1}]}^{consistent}$.
\end{itemize}

\subsubsection{3D Mask Preparation.} 

Similar to Stage 1, we project the consistent 2D masks $\mathcal{M}_t^\mathrm{consistent}$ from all frames $t \in [1, T]$ into 3D space using their corresponding depth information and camera parameters. However, unlike Stage 1 which only processes keyframes, Stage 2 extends the projection to all frames in the video sequence to leverage the temporally consistent masks generated by our tracking policy. So Stage 2's 3D point cloud set for each mask $M_t^{(j)}$ is obtained by:
\begin{equation}
    \mathcal{P}_t^{(j)} = \{(X_w, Y_w, Z_w) | (u,v) \in M_t^{(j)}, D_t(u,v) > 0\}.
\label{eq:stage2_point_set}
\end{equation}
where $(u,v)$ are pixel coordinates in the 2D mask, $(X_w, Y_w, Z_w)$ are the corresponding 3D world coordinates obtained through the same two-step transformation process defined in Eq.~\eqref{eq:pixel_to_camera} and Eq.~\eqref{eq:camera_to_world}. Note that $\mathcal{P}_t^{(j)}$ differs from $\mathcal{P}_{t_k}^{(j)}$ in Eq.~\eqref{eq:stage1_point_set} only in the frame indexing: while Stage 1 operates exclusively on keyframes $t_k$, Stage 2 processes keyframes and their surrounding temporal frames $t$ to fully utilize the cross-frame consistency established by our tracking approach.\par


The complete set of 3D pseudo-labels for Stage 2, denoted as $\tilde{Y}^{(2)}$, is the aggregation of all such projected point clouds:
\begin{equation}
    \tilde{Y}^{(2)} = \bigcup_{t=1}^{T} \bigcup_{j=1}^{N_t} \mathcal{P}_t^{(j)}.
\end{equation}

\subsubsection{Model Training.} 

Fine-tuning with Consistent Labels. We fine-tune the model $\text{Net}_1$ from Stage 1 using the granularity consistent pseudo-labels $\tilde{Y}^{(2)}$ to obtain the Stage 2 model, denoted as $\text{Net}_2$. The key distinction from Stage 1 is that $\tilde{Y}^{(2)}$ contains cross-frame consistent annotations that resolve segmentation granularity conflicts, enabling the model to learn robust cross-view correspondences and temporal relationships. The training objective follows the same formulation as Stage 1:
\begin{equation}
\begin{aligned}
\mathcal{L}_\mathrm{stage2} ={} & \sum_{p \in \tilde{Y}^{(2)}} \left[ \lambda_\mathrm{dice} \mathcal{L}_\mathrm{dice}(\text{Net}_2(p), \hat{y}_p) + \lambda_\mathrm{ce} \mathcal{L}_\mathrm{ce}(\text{Net}_2(p), \hat{y}_p) \right] \\
                       & + \lambda_\mathrm{obj} \mathcal{L}_\mathrm{obj}.
\end{aligned}
\end{equation}
where $\hat{y}_p$ represents the pseudo-label for point $p$ from $\tilde{Y}^{(2)}$ and $\text{Net}_2(p)$ is the corresponding prediction. 

\subsection{Stage 3: Full-Scene Fine-Tuning on 3D Point Clouds}
\label{sec:stage3}
The objective of Stage 3 is to enhance the understanding ability of the full 3D scene. We first use the Stage 2 model $\text{Net}_2$ to generate 3D pseudo labels $\tilde{Y}_\mathrm{Full}^{(3)} = \text{Net}_2(\mathcal{P}_\mathrm{full}) = \{{\hat{y}_i\}_{i=1}^{N}}$ on the full point cloud, where $\hat{y}_i \in \{1, 2, ..., C\}$ represents the pseudo label for the point $p_i$, with $C$ being the total number of object categories.
\subsubsection{Confidence-based Filtering.} To ensure the quality of pseudo labels, we apply confidence-based filtering: $\tilde{Y}_\mathrm{Full-filtered}^{(3)} = \{y_i | \max_c P(y_i = c | p_i) > \tau_{conf}\}$, where $P(y_i = c | p_i)$ is the probability that point $p_i$ belongs to category $c$ and $\tau_{conf}$ is the confidence threshold.

\subsubsection{Fine-tuning Objective.} The model is fine-tuned using the filtered pseudo-labels and the following loss:
\begin{equation}
\scalebox{0.9}{%
  $\displaystyle
  \mathcal{L}_\mathrm{stage3} = \sum_{p_i \in \tilde{Y}_\mathrm{Full-filtered}^{(3)}} \left[ \lambda_\mathrm{dice} \mathcal{L}_\mathrm{dice}(\text{Net}_{3}(p_i), y_i) + \lambda_\mathrm{ce} \mathcal{L}_\mathrm{ce}(\text{Net}_{3}(p_i), y_i) \right].
  $%
}
\end{equation}
where $\text{Net}_3$ is the final stage 3 model. This stage leverages the complete geometric structure of the scene to refine segmentation boundaries and resolve ambiguities that may exist in the projected 2D-to-3D pseudo-labels from previous stages.

\section{Experiments}

\begin{table*}[!t]
    \caption{Segmentation Score on ScanNet++~\cite{yeshwanth2023scannet++}. The metric is average precision (AP) on the validation split. We include fully-supervised Mask3D~\cite{schult2023mask3d} trained on manual ScanNet and ScanNet200 labels, and methods without ground truth labels.}
    \centering
    % ...
    \resizebox{\textwidth}{!}{%
    \begin{tabular}{llllll}
        \toprule
        Method & Ground Truth Labels & Avg. Inference Times/s
 &$AP$  &$AP_{50}$  &$AP_{25}$   \\
        \midrule
        \emph{Fully-supervised methods} &  &  &  &  & \\
        Mask3D~\cite{schult2023mask3d}  &ScanNet200  &0.7  & 8.7  & 15.5  &27.2  \\
        Mask3D~\cite{schult2023mask3d} &ScanNet  &0.7  & 9.4  & 16.8  &28.7  \\
        \midrule
        \emph{Without GT masks methods} &  &  &  &  & \\
        SAM3D~\cite{yang2023sam3d} & $\times$ & 386.7 & 3.9  & 9.3  & 22.1  \\
        Felzenszwalb et al.~\cite{felzenszwalb2004efficient}  & $\times$ &12.6  & 5.8  & 11.6  & 27.2   \\
        Segment3D~\cite{huang2024segment3d} & $\times$ &0.7  & 15.0  & 25.9  & 38.8   \\
        Ours  & $\times$ &0.7  & \textbf{17.7}   & \textbf{29.6}  & \textbf{42.5}   \\
        \bottomrule
    \end{tabular}
    }
    \label{tab:overall}
\end{table*}

\begin{table}
    \caption{Segmentation Score on ScanNet200~\cite{rozenberszki2022language}. The evaluation metric is average precision (AP) on the validation split.}
    \centering
    % ...
    \begin{tabular}{llll}
        \toprule
        Method       & $AP$  & $AP_{50}$  & $AP_{25}$   \\
        \midrule
        \emph{Fully-supervised methods} &  &  & \\
        Mask3D~\cite{schult2023mask3d}       & 34.1  & 43.1  &-  \\
        \midrule
        \emph{Without GT masks methods} &  &  & \\
        Felzenszwalb et al.~\cite{felzenszwalb2004efficient}  & 6.1  & 12.1  & -   \\
        UnScene3D~\cite{rozenberszki2024unscene3d} & 15.9  & 32.2  & -  \\
        SAM3D~\cite{yang2023sam3d} & 19.0  & 32.5  & -  \\
        Segment3D~\cite{huang2024segment3d}   & 27.0  & 39.1  & 50.3  \\
        Ours     & \textbf{30.2}  & \textbf{42.8}  & \textbf{52.5}  \\
        \bottomrule
    \end{tabular}
    \label{tab:results_on_scannet200}
\end{table}

\subsection{Basic Setups}
\subsubsection{Dataset}
To evaluate the effectiveness of our proposed method and conduct fair comparison with other SOTA methods, we adopt widely-used 3D instance segmentation datasets: ScanNet~\cite{dai2017scannet}, ScanNet200~\cite{rozenberszki2022language} and ScanNet++~\cite{yeshwanth2023scannet++}. ScanNet and ScanNet200 share the same indoor scenes, which contains 1,201 training scenes and 312 validation scenes, annotated with 200 object categories. ScanNet++ offers posed RGB-D images and sub-millimeter resolution 3D reconstructions, including 856 training scenes, 50 validation scenes and 50 test scenes, covering 1,659+ semantic and instance annotations. We train our method on ScanNet training set, evaluate on both ScanNet++ and ScanNet200 evaluation set. \par

\subsubsection{Metrics}
We report average precision(AP) scores at intersection over union (IoU) thresholds from 50\% to 95\%, in 25\% increments, and 50\%, 25\%, denoted as $AP_{50}$ and $AP_{25}$ to evaluate the class-agnostic segmentation results.\par

\subsubsection{Baseline}
We compare our approach with both full-supervised and without manual labels baselines. Mask3D~\cite{schult2023mask3d} is the state-of-the-art, transformer-based method, supervised with manually annotated 3D segmentation masks. Our method adopt the same backbone as Mask3D but instead of traning on manually annotated 3D masks, we learns from automatically generated masks. 

\subsection{Main Results}
\subsubsection{Results on ScanNet++ and ScanNet200} 
Our method is trained on the ScanNet training set without using any manual annotations. We evaluate its performance on both the ScanNet++ and the ScanNet200 validation set, with results presented in Tab.~\ref{tab:overall} and Tab.~\ref{tab:results_on_scannet200}, respectively. As shown in Tab.~\ref{tab:overall}, with an $AP$/$AP_{50}$/$AP_{25}$ of 17.7/29.6/42.5, outperforming the previous best method Segment3D~\citep{huang2024segment3d} (15.0) by 2.7/3.7/3.7 points, while maintaining comparable real-time inference speed of 0.7s per scene. This cross-dataset evaluation, where the model is trained on ScanNet but tested on ScanNet++, highlights the strong generalization capability of our method. Tab.~\ref{tab:results_on_scannet200} further demonstrates our model's effectiveness on the ScanNet200 validation set, achieving an AP of $AP$/$AP_{50}$/$AP_{25}$ of 30.2/42.8/52.5, and surpassing Segment3D by 3.2/3.7/2.2 points.


\begin{figure*}[!h]
    \centering
    \includegraphics[width=\textwidth]{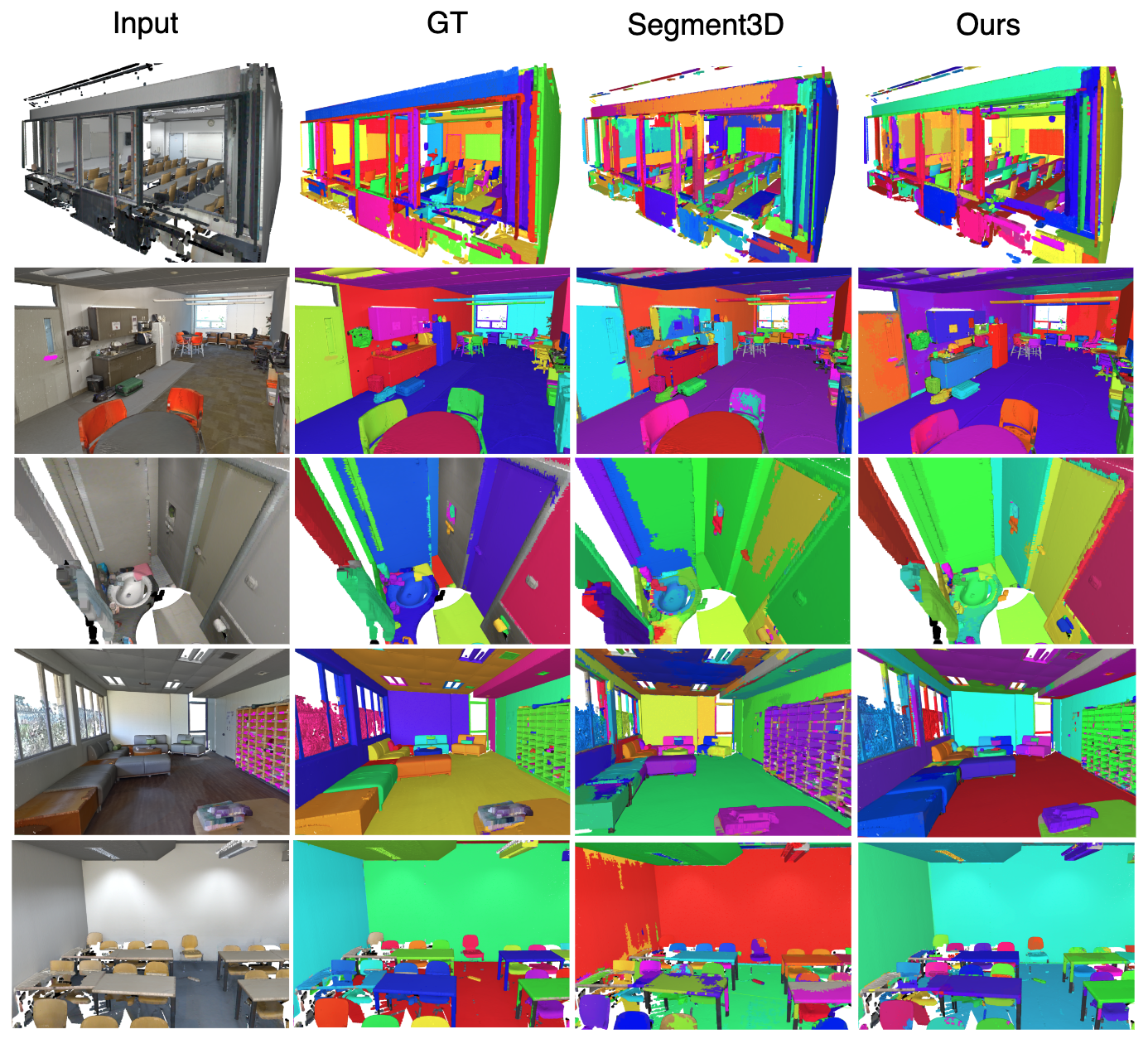}
    \caption{Qualitative Comparison of ScanNet++'s ground truth, Segment3D~\cite{huang2024segment3d} and ours.}
    \label{fig:quality_results}
\end{figure*}

\subsubsection{Qualitative Results}
Fig.~\ref{fig:quality_results} shows several representative comparison examples of segmentation results on the ScanNet++ dataset, across a variety of scenes (e.g., \textit{living room}, \textit{pantry}, \textit{classroom}, \textit{bathroom}) and instances, from the holistic to the focused perspective. Our method outperforms baseline method Segment3D~\cite{huang2024segment3d} in terms of segmentation integrity, leading to less noisy and more consistent results.

\subsection{Ablation and Analysis}

\subsubsection{The Importance of Consistency}

Table~\ref{tab:ablation_of_consistency} highlights the critical role of temporal granularity consistency. By progressively incorporating more intermediate frames with tracked, consistent masks for supervision, we observe a steady improvement in performance across all metrics compared to using only keyframes. This confirms that denser, more consistent supervision is key to enhancing the model's segmentation accuracy.

\begin{table}[!t]
    \caption{The importance of consistency. $K$ means only Key frames are adopted in the training, $K_{plus*}$ means plus additional $*$ frames supervision during training.}
    \centering
    % ...
    \begin{tabular}{llll}
        \toprule
        Method       & $AP$  & $AP_{50}$  & $AP_{25}$   \\
        \midrule
        \emph{ScanNet200}~\cite{rozenberszki2022language} &  &  & \\
        $K$      & 28.1   & 40.4  & 51.0   \\
        $K_{plus1}$ & \underline{29.7}  & \underline{42.3}  & \underline{52.2}  \\
        $K_{plus2}$      & \textbf{30.2} & \textbf{42.8}  & \textbf{52.5} \\
        \midrule
        \emph{ScanNet++}~\cite{yeshwanth2023scannet++} &  &  & \\
        $K$     & 16.3   & 28.0  & 40.3   \\
        $K_{plus1}$  & \underline{17.5} & \underline{29.6} & \underline{41.5}  \\
        $K_{plus2}$  & \textbf{17.7}  & \textbf{29.6}  & \textbf{42.5}    \\
        \bottomrule
    \end{tabular}
    \label{tab:ablation_of_consistency}
\end{table}

\subsubsection{The importance of Three-Stage Training}
We compare the performance of Segment3D pre-trained solely on partial RGB-D point clouds (Stage 1), then fine-tuning with consistency point clouds (Stage 2), finally fine-tuning on full point clouds (Stage 3). Scores are reported in Tab.~\ref{tab:importance_of_training_stage}, from which we can observe the importance of each stage. The results demonstrate that when all three stages are employed, the model achieves optimal performance. 
\begin{table}[!t]
    \caption{The importance of Three-Stage Training.}
    \centering
    % ...
    \begin{tabular}{lllllll}
        \toprule
        Stage 1  &Stage 2   &Stage 3  &$AP$  &$AP_{50}$  &$AP_{25}$   \\
        \midrule
         \checkmark  &    &              &11.4  &20.6  &34.6   \\
                 &\checkmark   &         &11.2  &20.4  &34.1   \\
         \checkmark  &    &\checkmark   &15.0 &25.9  &38.8   \\
           &\checkmark  &\checkmark  &14.9 &26.5 &39.9 \\
         \checkmark &\checkmark &\checkmark &\textbf{16.3} &\textbf{27.9} &\textbf{40.3}  \\
        \bottomrule
    \end{tabular}
    \label{tab:importance_of_training_stage}
\end{table}

\begin{table}[!t]
    \caption{The Performance of Open-Vocabulary Scene Understanding on Scannet200.}
    \centering
    \scalebox{0.85}{
    \begin{tabular}{lllllll}
        \toprule
        Segmentor &$AP$  &$AP_{50}$  &$AP_{25}$    &Head(AP)  &Common(AP)  &Tail(AP) \\
        \midrule
        \emph{Fully-supervised methods} &  &  & \\
        Mask3D  & 15.2  & 19.6  & 22.4 & 15.8  & 14.2  & 15.7 \\
        \midrule
        \emph{Zero-shot methods} &  &  & \\
        Segment3D  & 7.9  & 11.5  & 15.4 & 7.4  & 6.4  & 10.2 \\
        Ours  & 8.1  & 11.1  & 15.0 & 6.9 & 7.4  & 10.3 \\
        \bottomrule
    \end{tabular}
    }
    \label{tab:application_of_open_set_scene_understanding}
\end{table}

\subsection{Application: Open-Vocabulary Scene Understanding}
\label{sec:Open_voca}

Open-vocabulary scene understanding requires models to identify and localize objects based on arbitrary natural language queries, extending beyond predefined object categories. This task takes as input a 3D scene point cloud and a text query describing the target object, and outputs the corresponding 3D instance mask with similarity scores.\par
To extend our proposed method to this task, we associate the generated 3D masks with text features through a three-step pipeline: (1) Class-Agnostic Mask Computation: We use three different pre-trained models (fully-supervised Mask3D~\cite{schult2023mask3d}, Segment3D~\cite{huang2024segment3d}, and our method) to extract 3D instance masks from the scene; (2) Mask Feature Computation: For each 3D mask obtained from step (1), we project it onto multi-view RGB images and aggregate CLIP~\cite{radford2021learning-clip} visual features from the corresponding 2D regions to obtain semantic representations; (3) Text-Mask Association: We compute cosine similarity between the aggregated mask features and CLIP text embeddings of the query, ranking masks by similarity scores for retrieval.
We evaluate this application through two complementary approaches. In Sec.\ref{sec:Quantitative_analysis_open_voca}, we conduct quantitative evaluation on ScanNet200's 200 predefined categories, comparing the performance of different pretrained mask generation method (Mask3D, Segment3D and Ours), analyzing performance across Head, Common, and Tail object frequencies to demonstrate our method's effectiveness on standard benchmarks. While this evaluation reflects model's open-vocabulary scene understanding ability across different object classes, the ScanNet200 categories consist primarily of simple single-word vocabularies that cannot capture the full complexity of real-world object descriptions commonly encountered in practical applications. To explore our method's understanding of more diverse and nuanced 3D text queries, we further conduct qualitative analysis in Sec.\ref{sec:Qualitative_analysis_open_voca}, where we perform evaluation using diverse out-of-vocabulary queries with color, material, spatial, and functional descriptors, showcasing our method's superior fine-grained semantic understanding capabilities in real-world scenarios.

\subsubsection{Quantitative Analysis of Open-Vocabulary Scene Understanding}
\label{sec:Quantitative_analysis_open_voca}

As shown in Tab.~\ref{tab:application_of_open_set_scene_understanding}, we evaluate our model on ScanNet200, which categorizes its 200 classes into Head, Common, and Tail groups based on the frequency of labeled points in the training set. Our results show a distinct advantage in long-tail categories. The overall AP of 8.1 slightly outperforms Segment3D (7.9), primarily due to significant improvements in the Common (7.4 vs. 6.4) and Tail (10.3 vs. 10.2) categories. Notably, our model achieves its highest absolute performance on Tail categories (10.3), surpassing both Head (6.9) and Common (7.4) categories. This contradicts the conventional expectation that high-frequency categories should yield better performance and indicates that our zero-shot method excels at handling rare objects (e.g., \textit{`guitar'}, \textit{`clock'}, \textit{`stuffed animal'}), likely benefiting from the rich object representations in pre-trained vision-language models. In contrast, the relative underperformance on Head categories (6.9 vs. 7.4), which include large structural objects like \textit{`wall'} and \textit{`floor'}, along with a slight deficit in AP50/AP25 metrics, suggests that our method has room for improvement in the precise boundary localization of large-scale objects. This may be related to the inherent limitations of zero-shot methods when dealing with common indoor structures that require strict geometric constraints.

\begin{figure*}[!t]
    \centering
    \includegraphics[width=\textwidth]{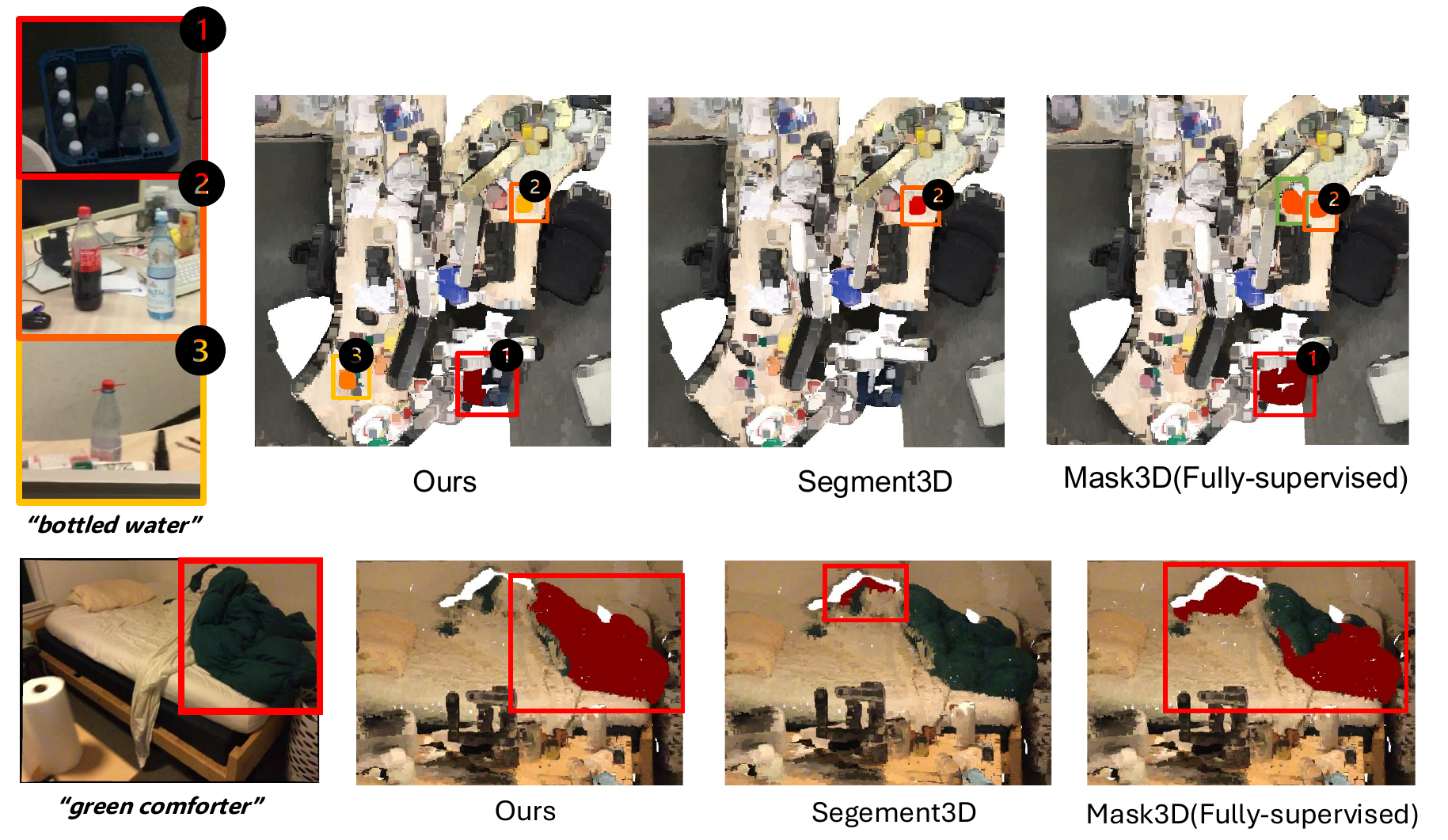}
    \caption{Comparison of 3D text retrieval results between our method, Segment3D, and the fully-supervised OpenMask3D. First row: Segmentation results for \textit{`bottled water'} query in an office scene containing three locations. Our method successfully identifies all instances, Segment3D only detects the second location, and OpenMask3D identifies the first two locations but misclassifies coca cola as bottled water. Second row: Segmentation results for \textit{`green comforter'} query in a bedroom scene, where our method achieves the most precise segmentation boundaries.}
    \label{fig:open_voca_comparison}
\end{figure*}

\subsubsection{Qualitative Analysis of Open-Vocabulary Scene Understanding}
\label{sec:Qualitative_analysis_open_voca}

To further validate our method's open-vocabulary capabilities, we evaluate 3D object retrieval performance using natural language queries. In the visualization results, we use a color map where red indicates high similarity scores, yellow indicates moderate scores, and green denotes low similarity.\par

\begin{figure*}[!t]
    \centering
    \includegraphics[width=\textwidth]{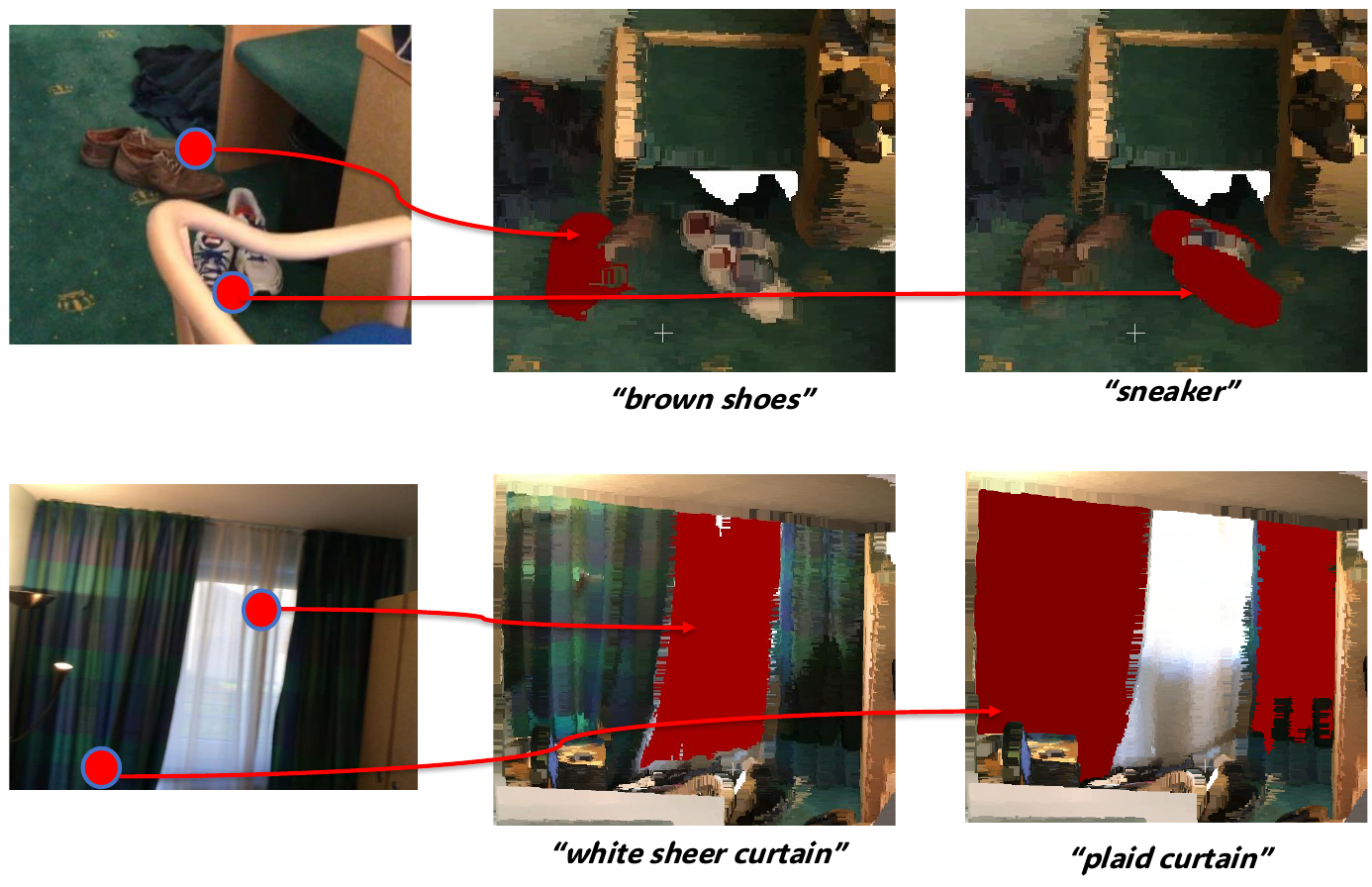}
    \caption{Our method's capability for recognizing similar object categories. First row: Accurate discrimination of shoe color and type. Second row: Effective identification of curtain color and material properties.}
    \label{fig:open_voca_similar}
\end{figure*}

\textbf{Fine-Grained Small Object Recognition.} Our method excels in fine-grained recognition of small objects that are prone to occlusion, even surpassing the supervised method, OpenMask3D. As shown in the first row of Fig.~\ref{fig:open_voca_comparison}, when retrieving \textit{`bottled water'} in an office scene containing three distinct locations with bottled water, our method accurately identifies all instances: the first location with the highest matching score, marked in deepest red, clear distinction from the adjacent \textit{`coca cola'} at the second location, and successful detection of the partially occluded bottles at the third location. In contrast, Segment3D fails to detect bottled water at the first and third locations, while OpenMask3D not only misses the third location but also cannot effectively distinguish between bottled water and coca cola at the second location, assigning identical color coding indicating equivalent matching scores, as highlighted by the green boxes. As demonstrated in the second row of Fig.~\ref{fig:open_voca_comparison}, our method achieves the most accurate segmentation when retrieving \textit{`green comforter'} in a bedroom scene, showcasing our capability to handle objects with irregular boundaries. \par

\textbf{Discriminative Recognition of Similar Objects.} 
Fig.~\ref{fig:open_voca_similar} illustrates our method's ability to distinguish between visually similar objects. In the upper row, our model accurately differentiates shoes based on both functional characteristics \textit{`sneaker'} and color features \textit{`brown shoes'}. The lower row demonstrates effective material and color discrimination for curtains; our method successfully distinguishes \textit{`white sheer curtain}' from \textit{`plaid curtain'}.\par

\begin{figure*}[!t]
    \centering
    \includegraphics[width=\textwidth]{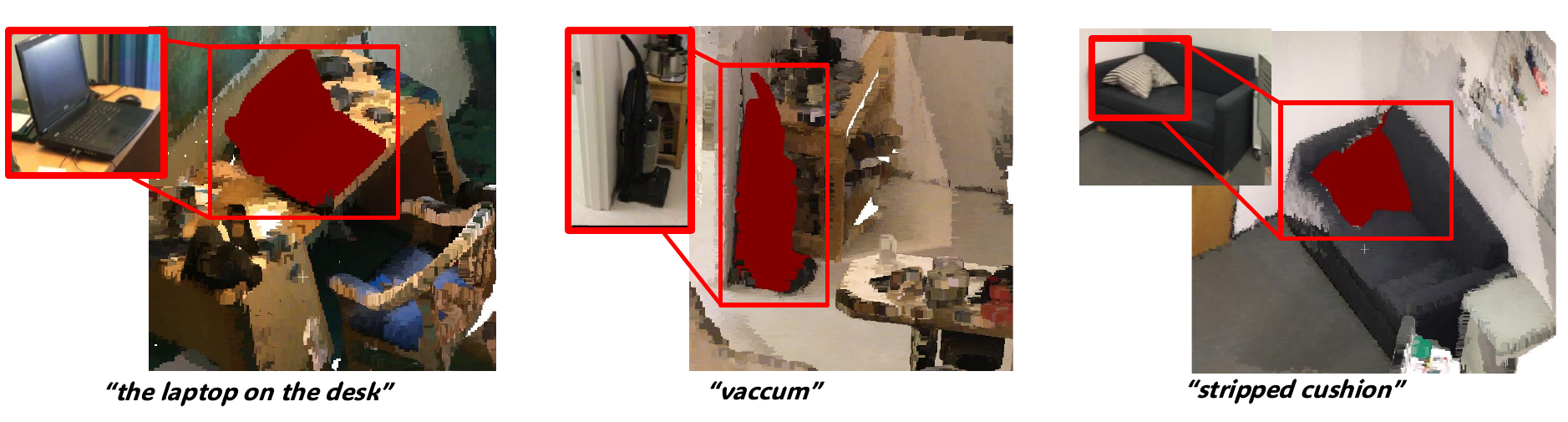}
    \caption{Our method's performance on long-tail categories. These categories have limited annotations in the dataset, yet our method successfully identifies and localizes them.}
    \label{fig:open_voca_long_tail}
\end{figure*}

\textbf{Long-tail Category Recognition.}
We further evaluate our method on long-tail categories not present in ScanNet200, as shown in Fig.~\ref{fig:open_voca_long_tail}. Our method successfully retrieves objects described with spatial relationships \textit{`the laptop on the desk'}, common household items absent from annotated datasets \textit{`vacuum'}, and attribute-based descriptions \textit{`striped cushion'}.

Overall, the aforementioned text queries encompass diverse categories with color, functional, material, and spatial descriptors that are intuitive and practical for real-world applications, yet are not included in ScanNet200's 200 predefined categories. These advantages extend beyond standard evaluation metrics and are not reflected in conventional AP assessments. While ScanNet200 contains 200 categories, most consist of simple single-word vocabularies that cannot capture the full complexity and fine-grained descriptions of real-world objects. In these ``out-of-vocabulary'' queries, our performance occasionally surpasses that of the supervised OpenMask3D method, further validating our approach's potential for real-world applications. We attribute this fine-grained semantic understanding capability to our 2D tracking mechanism, which maintains consistent object identification across multiple frames, while curriculum learning enhances the model's comprehension of spatial relationships and complex occlusion scenarios through progressive learning.

\section{Conclusion}

In this paper, we propose a Granularity-Consistent Segmentation Policy combined with a three-stage curriculum learning framework for class-agnostic 3D instance segmentation without manual labels. Our method addresses a critical limitation in existing training-based methods: the generation of inconsistent and conflicting 3D pseudo labels due to frame-independent 2D mask processing, which degrades segmentation quality when learning from 2D foundation models. Our Granularity-Consistent Segmentation Policy establishes temporal consistency across frames through automatic 2D mask tracking with object state management, producing unified pseudo-labels that eliminate cross-frame granularity conflicts. Building upon these consistent annotations, our three-stage curriculum learning framework progressively trains the model from fragmented keyframe annotations to consistent multi-view supervision, and finally to globally coherent full-scene understanding. This consistency-aware progressive training proved crucial for learning high-quality segmentations from initially fragmented and contradictory 2D priors. Our extensive experiments on the ScanNet200 and ScanNet++ benchmarks validate the effectiveness of our method, demonstrating state-of-the-art performance with real-time inference speed. The ablation studies further confirmed the importance of both our consistency policy and the multi-stage training pipeline. Furthermore, we demonstrated the practical value of our approach through open-vocabulary scene understanding experiments, showing superior performance in fine-grained object retrieval and long-tail category recognition, with particular strength in handling out-of-vocabulary queries that extend beyond predefined object categories. This capability makes our approach particularly suitable for real-world applications requiring flexible and intuitive human-robot interaction, establishing a foundation that enables diverse downstream 3D semantic understanding tasks.

\textbf{Acknowledgement.} This work was supported by JST SPRING, Grant Number JPMJSP2114, JSPS KAKENHI, Grant Numbers JP24H00733 and JP25K03130.






 \bibliographystyle{elsarticle-num-names} 
 \bibliography{ref}

\begin{thebibliography}{40}
\expandafter\ifx\csname natexlab\endcsname\relax\def\natexlab#1{#1}\fi
\providecommand{\url}[1]{\texttt{#1}}
\providecommand{\href}[2]{#2}
\providecommand{\path}[1]{#1}
\providecommand{\DOIprefix}{doi:}
\providecommand{\ArXivprefix}{arXiv:}
\providecommand{\URLprefix}{URL: }
\providecommand{\Pubmedprefix}{pmid:}
\providecommand{\doi}[1]{\href{http://dx.doi.org/#1}{\path{#1}}}
\providecommand{\Pubmed}[1]{\href{pmid:#1}{\path{#1}}}
\providecommand{\bibinfo}[2]{#2}
\ifx\xfnm\relax \def\xfnm[#1]{\unskip,\space#1}\fi
\bibitem[{Schult et~al.(2023)Schult, Engelmann, Hermans, Litany, Tang, and Leibe}]{schult2023mask3d}
\bibinfo{author}{J.~Schult}, \bibinfo{author}{F.~Engelmann}, \bibinfo{author}{A.~Hermans}, \bibinfo{author}{O.~Litany}, \bibinfo{author}{S.~Tang}, \bibinfo{author}{B.~Leibe},
\newblock \bibinfo{title}{{Mask3D}: Mask transformer for {3D} semantic instance segmentation},
\newblock in: \bibinfo{booktitle}{{International Conference on Robotics and Automation (ICRA)}}, \bibinfo{year}{2023}.
\bibitem[{Vu et~al.(2022)Vu, Kim, Luu, Nguyen, and Yoo}]{vu2022softgroup}
\bibinfo{author}{T.~Vu}, \bibinfo{author}{K.~Kim}, \bibinfo{author}{T.~M. Luu}, \bibinfo{author}{X.~T. Nguyen}, \bibinfo{author}{C.~D. Yoo},
\newblock \bibinfo{title}{{SoftGroup} for {3D} instance segmentation on {3D} point clouds},
\newblock in: \bibinfo{booktitle}{Proceedings of the IEEE Conference on Computer Vision and Pattern Recognition (CVPR)}, \bibinfo{year}{2022}.
\bibitem[{Dai et~al.(2017)Dai, Chang, Savva, Halber, Funkhouser, and Nie{\ss}ner}]{dai2017scannet}
\bibinfo{author}{A.~Dai}, \bibinfo{author}{A.~X. Chang}, \bibinfo{author}{M.~Savva}, \bibinfo{author}{M.~Halber}, \bibinfo{author}{T.~Funkhouser}, \bibinfo{author}{M.~Nie{\ss}ner},
\newblock \bibinfo{title}{{ScanNet}: Richly-annotated {3D} reconstructions of indoor scenes},
\newblock in: \bibinfo{booktitle}{Proceedings of the IEEE Conference on Computer Vision and Pattern Recognition (CVPR)}, \bibinfo{year}{2017}.
\bibitem[{Rozenberszki et~al.(2022)Rozenberszki, Litany, and Dai}]{rozenberszki2022language-scannet200}
\bibinfo{author}{D.~Rozenberszki}, \bibinfo{author}{O.~Litany}, \bibinfo{author}{A.~Dai},
\newblock \bibinfo{title}{Language-grounded indoor {3D} semantic segmentation in the wild},
\newblock in: \bibinfo{booktitle}{Proceedings of the European Conference on Computer Vision ({ECCV})}, \bibinfo{year}{2022}.
\bibitem[{Yeshwanth et~al.(2023)Yeshwanth, Liu, Nie{\ss}ner, and Dai}]{yeshwanth2023scannet++}
\bibinfo{author}{C.~Yeshwanth}, \bibinfo{author}{Y.-C. Liu}, \bibinfo{author}{M.~Nie{\ss}ner}, \bibinfo{author}{A.~Dai},
\newblock \bibinfo{title}{{ScanNet++}: A high-fidelity dataset of {3D} indoor scenes},
\newblock in: \bibinfo{booktitle}{Proceedings of the IEEE/CVF International Conference on Computer Vision (CVPR)}, \bibinfo{year}{2023}, pp. \bibinfo{pages}{12--22}.
\bibitem[{Huang et~al.(2024)Huang, Peng, Takmaz, Tombari, Pollefeys, Song, Huang, and Engelmann}]{huang2024segment3d}
\bibinfo{author}{R.~Huang}, \bibinfo{author}{S.~Peng}, \bibinfo{author}{A.~Takmaz}, \bibinfo{author}{F.~Tombari}, \bibinfo{author}{M.~Pollefeys}, \bibinfo{author}{S.~Song}, \bibinfo{author}{G.~Huang}, \bibinfo{author}{F.~Engelmann},
\newblock \bibinfo{title}{{Segment3D}: Learning fine-grained class-agnostic {3D} segmentation without manual labels},
\newblock in: \bibinfo{booktitle}{Proceedings of the European Conference on Computer Vision ({ECCV})}, \bibinfo{year}{2024}.
\bibitem[{Kirillov et~al.(2023)Kirillov, Mintun, Ravi, Mao, Rolland, Gustafson, Xiao, Whitehead, Berg, Lo et~al.}]{kirillov2023segment}
\bibinfo{author}{A.~Kirillov}, \bibinfo{author}{E.~Mintun}, \bibinfo{author}{N.~Ravi}, \bibinfo{author}{H.~Mao}, \bibinfo{author}{C.~Rolland}, \bibinfo{author}{L.~Gustafson}, \bibinfo{author}{T.~Xiao}, \bibinfo{author}{S.~Whitehead}, \bibinfo{author}{A.~C. Berg}, \bibinfo{author}{W.-Y. Lo}, et~al.,
\newblock \bibinfo{title}{Segment anything},
\newblock in: \bibinfo{booktitle}{Proceedings of the IEEE/CVF International Conference on Computer Vision (CVPR)}, \bibinfo{year}{2023}, pp. \bibinfo{pages}{4015--4026}.
\bibitem[{Lu et~al.(2023)Lu, Chang, Jing, Boularias, and Bekris}]{lu2023ovir}
\bibinfo{author}{S.~Lu}, \bibinfo{author}{H.~Chang}, \bibinfo{author}{E.~P. Jing}, \bibinfo{author}{A.~Boularias}, \bibinfo{author}{K.~Bekris},
\newblock \bibinfo{title}{{OVIR-3D}: Open-vocabulary {3D} instance retrieval without training on 3d data},
\newblock in: \bibinfo{booktitle}{7th Annual Conference on Robot Learning}, \bibinfo{year}{2023}.
\bibitem[{Yunhan~Yang et~al.(2023)Yunhan~Yang, Zhao, and Liu}]{yang2023sam3d}
\bibinfo{author}{T.~H. Yunhan~Yang, Xiaoyang~Wu}, \bibinfo{author}{H.~Zhao}, \bibinfo{author}{X.~Liu},
\newblock \bibinfo{title}{{SAM3D}: Segment anything in {3D} scenes},
\newblock in: \bibinfo{booktitle}{Proceedings of the IEEE/CVF International Conference on Computer Vision (ICCV) Workshops}, \bibinfo{year}{2023}.
\bibitem[{Yin et~al.(2024)Yin, Liu, Xiao, Cohen-Or, Huang, and Chen}]{yin2024sai3d}
\bibinfo{author}{Y.~Yin}, \bibinfo{author}{Y.~Liu}, \bibinfo{author}{Y.~Xiao}, \bibinfo{author}{D.~Cohen-Or}, \bibinfo{author}{J.~Huang}, \bibinfo{author}{B.~Chen},
\newblock \bibinfo{title}{{SAI3D}: Segment any instance in {3D} scenes},
\newblock in: \bibinfo{booktitle}{Proceedings of the IEEE/CVF Conference on Computer Vision and Pattern Recognition (CVPR)}, \bibinfo{year}{2024}, pp. \bibinfo{pages}{3292--3302}.
\bibitem[{Lu et~al.(2023)Lu, Kuen, Tiancheng, Jiuxiang, Weidong, Jiaya, Zhe, and Ming-Hsuan}]{qi2022high-cropformer}
\bibinfo{author}{Q.~Lu}, \bibinfo{author}{J.~Kuen}, \bibinfo{author}{S.~Tiancheng}, \bibinfo{author}{G.~Jiuxiang}, \bibinfo{author}{G.~Weidong}, \bibinfo{author}{J.~Jiaya}, \bibinfo{author}{L.~Zhe}, \bibinfo{author}{Y.~Ming-Hsuan},
\newblock \bibinfo{title}{High-quality entity segmentation},
\newblock in: \bibinfo{booktitle}{Proceedings of the IEEE/CVF International Conference on Computer Vision (ICCV)}, \bibinfo{year}{2023}.
\bibitem[{Guo et~al.(2024)Guo, Zhu, Peng, Wang, Shen, Hu, and Zhou}]{guo2024sam-graph}
\bibinfo{author}{H.~Guo}, \bibinfo{author}{H.~Zhu}, \bibinfo{author}{S.~Peng}, \bibinfo{author}{Y.~Wang}, \bibinfo{author}{Y.~Shen}, \bibinfo{author}{R.~Hu}, \bibinfo{author}{X.~Zhou},
\newblock \bibinfo{title}{{SAM}-guided graph cut for {3D} instance segmentation},
\newblock in: \bibinfo{booktitle}{Proceedings of the European Conference on Computer Vision ({ECCV})}, \bibinfo{year}{2024}, pp. \bibinfo{pages}{234--251}.
\bibitem[{Rozenberszki et~al.(2022)Rozenberszki, Litany, and Dai}]{rozenberszki2022language}
\bibinfo{author}{D.~Rozenberszki}, \bibinfo{author}{O.~Litany}, \bibinfo{author}{A.~Dai},
\newblock \bibinfo{title}{{Language-Grounded Indoor 3D Semantic Segmentation in the Wild}},
\newblock in: \bibinfo{booktitle}{Proceedings of the European Conference on Computer Vision ({ECCV})}, \bibinfo{year}{2022}.
\bibitem[{Wang et~al.(2024)Wang, Wang, Zhang, Guo, Liu, and Wang}]{wang2023openinst}
\bibinfo{author}{C.~Wang}, \bibinfo{author}{G.~Wang}, \bibinfo{author}{Q.~Zhang}, \bibinfo{author}{P.~Guo}, \bibinfo{author}{W.~Liu}, \bibinfo{author}{X.~Wang},
\newblock \bibinfo{title}{{OpenInst: A Simple Query-Based Method for Open-World Instance Segmentation}},
\newblock \bibinfo{journal}{Pattern Recognition} \bibinfo{volume}{153} (\bibinfo{year}{2024}) \bibinfo{pages}{110570}.
\bibitem[{Qi et~al.(2022)Qi, Kuen, Wang, Gu, Zhao, Torr, Lin, and Jia}]{qi2022open}
\bibinfo{author}{L.~Qi}, \bibinfo{author}{J.~Kuen}, \bibinfo{author}{Y.~Wang}, \bibinfo{author}{J.~Gu}, \bibinfo{author}{H.~Zhao}, \bibinfo{author}{P.~Torr}, \bibinfo{author}{Z.~Lin}, \bibinfo{author}{J.~Jia},
\newblock \bibinfo{title}{{Open world entity segmentation}},
\newblock \bibinfo{journal}{IEEE Transactions on Pattern Analysis and Machine Intelligence({TPAMI})}  (\bibinfo{year}{2022}).
\bibitem[{Tiancheng~Shen et~al.(2022)Tiancheng~Shen, Qi, Kuen, Xie, Wu, Lin, and Jia}]{shen2021high}
\bibinfo{author}{Y.~Z. Tiancheng~Shen}, \bibinfo{author}{L.~Qi}, \bibinfo{author}{J.~Kuen}, \bibinfo{author}{X.~Xie}, \bibinfo{author}{J.~Wu}, \bibinfo{author}{Z.~Lin}, \bibinfo{author}{J.~Jia},
\newblock \bibinfo{title}{{High Quality Segmentation for Ultra High-resolution Images}},
\newblock in: \bibinfo{booktitle}{Proceedings of the IEEE/CVF International Conference on Computer Vision (CVPR)}, \bibinfo{year}{2022}.
\bibitem[{Qi et~al.(2022)Qi, Kuen, Lin, Gu, Rao, Li, Guo, Wen, Yang, and Jia}]{qi2022cassl}
\bibinfo{author}{L.~Qi}, \bibinfo{author}{J.~Kuen}, \bibinfo{author}{Z.~Lin}, \bibinfo{author}{J.~Gu}, \bibinfo{author}{F.~Rao}, \bibinfo{author}{D.~Li}, \bibinfo{author}{W.~Guo}, \bibinfo{author}{Z.~Wen}, \bibinfo{author}{M.-H. Yang}, \bibinfo{author}{J.~Jia},
\newblock \bibinfo{title}{{CA-SSL: Class-Agnostic Semi-Supervised Learning for Detection and Segmentation}},
\newblock in: \bibinfo{booktitle}{Proceedings of the European Conference on Computer Vision ({ECCV})}, \bibinfo{year}{2022}.
\bibitem[{Yang et~al.(2019)Yang, Wang, Clark, Hu, Wang, Markham, and Trigoni}]{yang2019learning}
\bibinfo{author}{B.~Yang}, \bibinfo{author}{J.~Wang}, \bibinfo{author}{R.~Clark}, \bibinfo{author}{Q.~Hu}, \bibinfo{author}{S.~Wang}, \bibinfo{author}{A.~Markham}, \bibinfo{author}{N.~Trigoni},
\newblock \bibinfo{title}{Learning object bounding boxes for {3D} instance segmentation on point clouds},
\newblock \bibinfo{journal}{Advances in neural information processing systems (NeurIPS)} \bibinfo{volume}{32} (\bibinfo{year}{2019}).
\bibitem[{Yi et~al.(2019)Yi, Zhao, Wang, Sung, and Guibas}]{yi2019gspn}
\bibinfo{author}{L.~Yi}, \bibinfo{author}{W.~Zhao}, \bibinfo{author}{H.~Wang}, \bibinfo{author}{M.~Sung}, \bibinfo{author}{L.~J. Guibas},
\newblock \bibinfo{title}{{GSPN: Generative shape proposal network for 3D instance segmentation in point cloud}},
\newblock in: \bibinfo{booktitle}{Proceedings of the IEEE/CVF Conference on Computer Vision and Pattern Recognition (CVPR)}, \bibinfo{year}{2019}, pp. \bibinfo{pages}{3947--3956}.
\bibitem[{Chen et~al.(2021)Chen, Fang, Zhang, Liu, and Wang}]{chen2021hierarchical}
\bibinfo{author}{S.~Chen}, \bibinfo{author}{J.~Fang}, \bibinfo{author}{Q.~Zhang}, \bibinfo{author}{W.~Liu}, \bibinfo{author}{X.~Wang},
\newblock \bibinfo{title}{Hierarchical aggregation for {3D} instance segmentation},
\newblock in: \bibinfo{booktitle}{Proceedings of the IEEE/CVF International Conference on Computer Vision (ICCV)}, \bibinfo{year}{2021}, pp. \bibinfo{pages}{15467--15476}.
\bibitem[{Jiang et~al.(2020)Jiang, Zhao, Shi, Liu, Fu, and Jia}]{jiang2020pointgroup}
\bibinfo{author}{L.~Jiang}, \bibinfo{author}{H.~Zhao}, \bibinfo{author}{S.~Shi}, \bibinfo{author}{S.~Liu}, \bibinfo{author}{C.-W. Fu}, \bibinfo{author}{J.~Jia},
\newblock \bibinfo{title}{{PointGroup}: Dual-set point grouping for {3D} instance segmentation},
\newblock in: \bibinfo{booktitle}{Proceedings of the IEEE Conference on Computer Vision and Pattern Recognition (CVPR)}, \bibinfo{year}{2020}.
\bibitem[{Liang et~al.(2021)Liang, Li, Xu, Tan, and Jia}]{liang2021instance}
\bibinfo{author}{Z.~Liang}, \bibinfo{author}{Z.~Li}, \bibinfo{author}{S.~Xu}, \bibinfo{author}{M.~Tan}, \bibinfo{author}{K.~Jia},
\newblock \bibinfo{title}{Instance segmentation in {3D} scenes using semantic superpoint tree networks},
\newblock in: \bibinfo{booktitle}{Proceedings of the IEEE/CVF International Conference on Computer Vision (ICCV)}, \bibinfo{year}{2021}, pp. \bibinfo{pages}{2783--2792}.
\bibitem[{Lu et~al.(2023)Lu, Deng, Wang, He, and Zhang}]{lu2023query}
\bibinfo{author}{J.~Lu}, \bibinfo{author}{J.~Deng}, \bibinfo{author}{C.~Wang}, \bibinfo{author}{J.~He}, \bibinfo{author}{T.~Zhang},
\newblock \bibinfo{title}{Query refinement transformer for {3D} instance segmentation},
\newblock in: \bibinfo{booktitle}{Proceedings of the IEEE/CVF International Conference on Computer Vision (ICCV)}, \bibinfo{year}{2023}, pp. \bibinfo{pages}{18516--18526}.
\bibitem[{Wu et~al.(2024)Wu, Tian, Wen, Peng, Liu, Yu, and Zhao}]{wu2024ppt}
\bibinfo{author}{X.~Wu}, \bibinfo{author}{Z.~Tian}, \bibinfo{author}{X.~Wen}, \bibinfo{author}{B.~Peng}, \bibinfo{author}{X.~Liu}, \bibinfo{author}{K.~Yu}, \bibinfo{author}{H.~Zhao},
\newblock \bibinfo{title}{{Towards Large-scale 3D Representation Learning with Multi-dataset Point Prompt Training}},
\newblock in: \bibinfo{booktitle}{Proceedings of the IEEE/CVF International Conference on Computer Vision (CVPR)}, \bibinfo{year}{2024}.
\bibitem[{Wu et~al.(2022)Wu, Lao, Jiang, Liu, and Zhao}]{wu2022ptv2}
\bibinfo{author}{X.~Wu}, \bibinfo{author}{Y.~Lao}, \bibinfo{author}{L.~Jiang}, \bibinfo{author}{X.~Liu}, \bibinfo{author}{H.~Zhao},
\newblock \bibinfo{title}{{Point transformer V2: Grouped Vector Attention and Partition-based Pooling}},
\newblock \bibinfo{journal}{Advances in neural information processing systems (NeurIPS)}  (\bibinfo{year}{2022}).
\bibitem[{Wu et~al.(2024)Wu, Jiang, Wang, Liu, Liu, Qiao, Ouyang, He, and Zhao}]{wu2024ptv3}
\bibinfo{author}{X.~Wu}, \bibinfo{author}{L.~Jiang}, \bibinfo{author}{P.-S. Wang}, \bibinfo{author}{Z.~Liu}, \bibinfo{author}{X.~Liu}, \bibinfo{author}{Y.~Qiao}, \bibinfo{author}{W.~Ouyang}, \bibinfo{author}{T.~He}, \bibinfo{author}{H.~Zhao},
\newblock \bibinfo{title}{{Point Transformer V3: Simpler, Faster, Stronger}},
\newblock in: \bibinfo{booktitle}{Proceedings of the IEEE/CVF International Conference on Computer Vision (CVPR)}, \bibinfo{year}{2024}.
\bibitem[{Wu et~al.(2025)Wu, Sun, Xu, Jiang, Ma, and Zhang}]{wu2025class}
\bibinfo{author}{J.~Wu}, \bibinfo{author}{M.~Sun}, \bibinfo{author}{H.~Xu}, \bibinfo{author}{C.~Jiang}, \bibinfo{author}{W.~Ma}, \bibinfo{author}{Q.~Zhang},
\newblock \bibinfo{title}{Class agnostic and specific consistency learning for weakly-supervised point cloud semantic segmentation},
\newblock \bibinfo{journal}{Pattern Recognition} \bibinfo{volume}{158} (\bibinfo{year}{2025}) \bibinfo{pages}{111067}.
\bibitem[{Ravi et~al.(2024)Ravi, Gabeur, Hu, Hu, Ryali, Ma, Khedr, R{\"a}dle, Rolland, Gustafson, Mintun, Pan, Alwala, Carion, Wu, Girshick, Doll{\'a}r, and Feichtenhofer}]{ravi2024sam2}
\bibinfo{author}{N.~Ravi}, \bibinfo{author}{V.~Gabeur}, \bibinfo{author}{Y.-T. Hu}, \bibinfo{author}{R.~Hu}, \bibinfo{author}{C.~Ryali}, \bibinfo{author}{T.~Ma}, \bibinfo{author}{H.~Khedr}, \bibinfo{author}{R.~R{\"a}dle}, \bibinfo{author}{C.~Rolland}, \bibinfo{author}{L.~Gustafson}, \bibinfo{author}{E.~Mintun}, \bibinfo{author}{J.~Pan}, \bibinfo{author}{K.~V. Alwala}, \bibinfo{author}{N.~Carion}, \bibinfo{author}{C.-Y. Wu}, \bibinfo{author}{R.~Girshick}, \bibinfo{author}{P.~Doll{\'a}r}, \bibinfo{author}{C.~Feichtenhofer},
\newblock \bibinfo{title}{{SAM2}: Segment anything in images and videos},
\newblock \bibinfo{journal}{arXiv preprint arXiv:2408.00714}  (\bibinfo{year}{2024}).
\bibitem[{Zhang et~al.(2023)Zhang, Dong, and Ma}]{zhang2023clipfo3d}
\bibinfo{author}{J.~Zhang}, \bibinfo{author}{R.~Dong}, \bibinfo{author}{K.~Ma},
\newblock \bibinfo{title}{{CLIP-FO3D: Learning Free Open-world 3D Scene Representations from 2D Dense CLIP}},
\newblock in: \bibinfo{booktitle}{Proceedings of the IEEE/CVF international conference on computer vision (CVPR)}, \bibinfo{year}{2023}, pp. \bibinfo{pages}{2048--2059}.
\bibitem[{Ha and Song(2022)}]{ha2022semabs}
\bibinfo{author}{H.~Ha}, \bibinfo{author}{S.~Song},
\newblock \bibinfo{title}{{Semantic Abstraction: Open-World 3D Scene Understanding from 2D Vision-Language Models}},
\newblock in: \bibinfo{booktitle}{Proceedings of the 2022 Conference on Robot Learning (CoRL)}, \bibinfo{year}{2022}.
\bibitem[{Sun et~al.(2025)Sun, Xu, Fan, Lu, and Liu}]{sun2025ov}
\bibinfo{author}{Z.~Sun}, \bibinfo{author}{X.~Xu}, \bibinfo{author}{B.~Fan}, \bibinfo{author}{J.~Lu}, \bibinfo{author}{H.~Liu},
\newblock \bibinfo{title}{{OV-GT3D: A generalizable open-vocabulary two-stage 3D detector with dual path distillation}},
\newblock \bibinfo{journal}{Pattern Recognition}  (\bibinfo{year}{2025}) \bibinfo{pages}{112156}.
\bibitem[{Zhang et~al.(2025)Zhang, Gao, Ye, Jin, Jiang, and Yang}]{zhang2025clip-pr}
\bibinfo{author}{Z.~Zhang}, \bibinfo{author}{B.~Gao}, \bibinfo{author}{J.~Ye}, \bibinfo{author}{H.~Jin}, \bibinfo{author}{L.~Jiang}, \bibinfo{author}{W.~Yang},
\newblock \bibinfo{title}{{CLIP prior-guided 3D open-vocabulary occupancy prediction}},
\newblock \bibinfo{journal}{Pattern Recognition} \bibinfo{volume}{162} (\bibinfo{year}{2025}) \bibinfo{pages}{111347}.
\bibitem[{Radford et~al.(2021)Radford, Kim, Hallacy, Ramesh, Goh, Agarwal, Sastry, Askell, Mishkin, Clark et~al.}]{radford2021learning-clip}
\bibinfo{author}{A.~Radford}, \bibinfo{author}{J.~W. Kim}, \bibinfo{author}{C.~Hallacy}, \bibinfo{author}{A.~Ramesh}, \bibinfo{author}{G.~Goh}, \bibinfo{author}{S.~Agarwal}, \bibinfo{author}{G.~Sastry}, \bibinfo{author}{A.~Askell}, \bibinfo{author}{P.~Mishkin}, \bibinfo{author}{J.~Clark}, et~al.,
\newblock \bibinfo{title}{Learning transferable visual models from natural language supervision},
\newblock in: \bibinfo{booktitle}{Proceedings of the International Conference on Machine Learning (ICML)}, \bibinfo{year}{2021}, pp. \bibinfo{pages}{8748--8763}.
\bibitem[{Ding et~al.(2023)Ding, Yang, Xue, Zhang, Bai, and Qi}]{ding2022language-pla}
\bibinfo{author}{R.~Ding}, \bibinfo{author}{J.~Yang}, \bibinfo{author}{C.~Xue}, \bibinfo{author}{W.~Zhang}, \bibinfo{author}{S.~Bai}, \bibinfo{author}{X.~Qi},
\newblock \bibinfo{title}{{PLA: Language-Driven Open-Vocabulary {3D} Scene Understanding}},
\newblock in: \bibinfo{booktitle}{Proceedings of the IEEE/CVF Conference on Computer Vision and Pattern Recognition (CVPR)}, \bibinfo{year}{2023}.
\bibitem[{Yang et~al.(2024)Yang, Ding, Deng, Wang, and Qi}]{yang2024regionplc}
\bibinfo{author}{J.~Yang}, \bibinfo{author}{R.~Ding}, \bibinfo{author}{W.~Deng}, \bibinfo{author}{Z.~Wang}, \bibinfo{author}{X.~Qi},
\newblock \bibinfo{title}{{RegionPLC}: Regional point-language contrastive learning for open-world {3D} scene understanding},
\newblock in: \bibinfo{booktitle}{Proceedings of the IEEE/CVF Conference on Computer Vision and Pattern Recognition (CVPR)}, \bibinfo{year}{2024}.
\bibitem[{Peng et~al.(2023)Peng, Genova, Jiang, Tagliasacchi, Pollefeys, and Funkhouser}]{Peng2023OpenScene}
\bibinfo{author}{S.~Peng}, \bibinfo{author}{K.~Genova}, \bibinfo{author}{C.~M. Jiang}, \bibinfo{author}{A.~Tagliasacchi}, \bibinfo{author}{M.~Pollefeys}, \bibinfo{author}{T.~Funkhouser},
\newblock \bibinfo{title}{{OpenScene}: {3D} scene understanding with open vocabularies},
\newblock in: \bibinfo{booktitle}{Proceedings of the IEEE/CVF Conference on Computer Vision and Pattern Recognition (CVPR)}, \bibinfo{year}{2023}.
\bibitem[{Takmaz et~al.(2023)Takmaz, Fedele, Sumner, Pollefeys, Tombari, and Engelmann}]{takmaz2023openmask3d}
\bibinfo{author}{A.~Takmaz}, \bibinfo{author}{E.~Fedele}, \bibinfo{author}{R.~W. Sumner}, \bibinfo{author}{M.~Pollefeys}, \bibinfo{author}{F.~Tombari}, \bibinfo{author}{F.~Engelmann},
\newblock \bibinfo{title}{{OpenMask3D}: Open-vocabulary {3D} instance segmentation},
\newblock in: \bibinfo{booktitle}{Advances in Neural Information Processing Systems (NeurIPS)}, \bibinfo{year}{2023}.
\bibitem[{Choy et~al.(2019)Choy, Gwak, and Savarese}]{choy20194d}
\bibinfo{author}{C.~Choy}, \bibinfo{author}{J.~Gwak}, \bibinfo{author}{S.~Savarese},
\newblock \bibinfo{title}{{4D} spatio-temporal convnets: Minkowski convolutional neural networks},
\newblock in: \bibinfo{booktitle}{Proceedings of the IEEE Conference on Computer Vision and Pattern Recognition (CVPR)}, \bibinfo{year}{2019}, pp. \bibinfo{pages}{3075--3084}.
\bibitem[{Felzenszwalb and Huttenlocher(2004)}]{felzenszwalb2004efficient}
\bibinfo{author}{P.~F. Felzenszwalb}, \bibinfo{author}{D.~P. Huttenlocher},
\newblock \bibinfo{title}{Efficient graph-based image segmentation},
\newblock \bibinfo{journal}{International journal of computer vision} \bibinfo{volume}{59} (\bibinfo{year}{2004}) \bibinfo{pages}{167--181}.
\bibitem[{Rozenberszki et~al.(2024)Rozenberszki, Litany, and Dai}]{rozenberszki2024unscene3d}
\bibinfo{author}{D.~Rozenberszki}, \bibinfo{author}{O.~Litany}, \bibinfo{author}{A.~Dai},
\newblock \bibinfo{title}{{UnScene3D}: Unsupervised {3D} instance segmentation for indoor scenes},
\newblock in: \bibinfo{booktitle}{Proceedings of the IEEE/CVF Conference on Computer Vision and Pattern Recognition (CVPR)}, \bibinfo{year}{2024}, pp. \bibinfo{pages}{19957--19967}.

\end{thebibliography}







\end{document}